\documentclass[11pt,letterpaper]{article}

\usepackage[T1]{fontenc}
\usepackage[utf8]{inputenc}
\usepackage{lmodern}
\input{glyphtounicode}
\usepackage[margin=1in]{geometry}
\usepackage{microtype}
\usepackage{authblk}

\usepackage{amsmath,amssymb}
\usepackage{booktabs}
\usepackage{caption}
\usepackage{graphicx}
\usepackage{algorithm}
\usepackage{needspace}
\usepackage{algpseudocode}
\usepackage{flafter}
\usepackage{float}
\usepackage{placeins}
\usepackage{xcolor}
\usepackage[authoryear,round]{natbib}

\usepackage{xurl}
\usepackage[hidelinks]{hyperref}
\hypersetup{
  pdftitle={RPA: Residual Patch-Token Adapter for Image Retrieval from EEG and MEG},
  pdfsubject={Research preprint},
  pdfauthor={Yuhui Jin, Yonghao Song, Bingchuan Liu},
  pdfkeywords={EEG, MEG, image retrieval, patch tokens, residual patch adapter}
}

\newenvironment{breakablealgorithm}{%
  \par\addvspace{\intextsep}\Needspace{8\baselineskip}%
  \begingroup\captionsetup{type=algorithm,position=top,skip=4pt}%
  \hrule height .8pt\relax\nobreak\vspace{3pt}\nobreak
}{%
  \par\nobreak\vspace{3pt}\nobreak\hrule height .8pt\relax
  \endgroup\par\addvspace{\intextsep}%
}

\graphicspath{{figures/}}
\title{\bfseries RPA: Residual Patch-Token Adapter for Image Retrieval from EEG and MEG}

\author[1]{Yuhui Jin}
\author[2]{Yonghao Song}
\author[3]{Bingchuan Liu}
\affil[1]{Vortek Lab Inc.}
\affil[2]{Department of Computer Science and Technology, Tsinghua University}
\affil[3]{Biology and Biological Engineering, California Institute of Technology}

\date{}         

\newif\ifwithsupplement
\withsupplementtrue

\begin{document}
\maketitle
\begin{abstract}
Visual decoding from non-invasive MEG and EEG (M/EEG) aims to identify viewed images through representation learning that aligns brain signals with corresponding images. Most existing MEG and EEG (M/EEG) visual decoding methods align brain signals with a single global embedding extracted from a pretrained visual encoder, leaving open whether intermediate patch representations, which preserve richer and more granular rich visual information, can improve representation learning. To address this question, we introduce the Residual Patch Adapter (RPA), a lightweight, modular adapter that leverages all patch tokens from an intermediate layer of a ViT visual encoder for alignment. Through extensive ablation analyses, we first show that pooling or masking patch tokens degrades the learned representation, demonstrating that retaining the full set of patch tokens is important for EEG alignment, while the CLS token provides little unique information. We then use a series of six quantitative feature analyses to show that both higher-level semantics and lower-level visual features, including color and texture, are essential for this EEG-to-image alignment. Under current protocols, our system achieves Top-1 accuracies of 95.4\% within-subject and 35.5\% cross-subject on THINGS-EEG2, and 65.2\% and 6.7\%, respectively, on THINGS-MEG, achieving state-of-the-art (SOTA) performance across both datasets. Evaluations with alternative brain encoders, including pretrained EEG foundation models, demonstrate that the approach extends beyond the projection-based EEG encoder. Furthermore, we provide a plug-and-play interface that allows RPA to be replaced by convolution, attention, or ConvNeXt alternatives. Together, these findings provide significant insight into M/EEG-to-image representation learning by establishing design principles for leveraging the latent space of visual encoders, and open new directions for brain--image alignment and non-invasive brain--computer interface (BCI).
\end{abstract}

\section{Introduction}
\label{sec:introduction}

Non-invasive visual decoding aims to identify viewed images from EEG or MEG
responses. Contrastive approaches learn a shared embedding space in which a
brain response retrieves its corresponding image from a held-out gallery
\citep{song2024nice,li2024atm}. The visual embedding plays a central role in
this framework: it provides the image-side target during training and
represents each candidate image during retrieval. How this embedding is
constructed therefore shapes the visual information that the brain encoder
learns to recover.

Most existing systems ultimately represent each image with a single global
embedding from a pretrained visual encoder. Recent work has enriched this
image-side representation by transforming the stimulus, combining
complementary visual targets, fusing multiple visual encoders, or adapting
visual features to brain signals
\citep{wu2025ubp,liu2026blur,zhang2026neurobridge,wang2025vieeg,
jo2026hyfi,wu2026shrinking,zheng2026brainhive}. Other studies show that
intermediate visual-encoder layers, patch aggregation, and combinations across
layers can improve alignment
\citep{du2026shallow,tang2026aligning,samga2026}. However it remains unclear
whether intermediate patch tokens should be compressed into a fixed global
embedding or retained for learned adaptation. We therefore ask whether learning
from the complete patch-token set can produce a more effective visual
embedding for M/EEG-to-image alignment.

To address this question, we introduce the \emph{Residual Patch Adapter} (RPA),
a lightweight, modular visual adapter applied to all patch tokens from a
selected intermediate layer of a frozen visual encoder. RPA transforms the
complete token set into a single visual embedding. A
plug-and-play interface also allows RPA to be replaced by alternative visual
adapters without changing the visual encoder or the brain--image alignment
objective.

Our contributions are summarized as follows:
\begin{itemize}
    \item We introduce the Residual Patch Adapter (RPA), a lightweight, modular
    visual adapter that processes all patch tokens from a selected
    intermediate layer before global pooling while keeping the
    pretrained visual encoder frozen. We further provide a plug-and-play
    interface for fixed-pooling, conv-based, attention-based,
    and ConvNeXt visual adapters, allowing the adapter to be changed
    without modifying the frozen visual encoder or alignment objective; RPA
    performs best in this comparison.

    \item Within the selected visual-encoder output, we find that visual
    information recoverable from EEG is distributed across multiple patch
    tokens and is not captured as effectively by the CLS token or fixed
    patch mean.
    This evidence empirically favors passing the patch tokens through an
    adaptive network before global pooling.

    \item Our system achieves $95.4\%$ within-subject and $35.5\%$
leave-one-subject-out(LOSO) Top-1 accuracy on THINGS-EEG2, and $65.2\%$
$6.7\%$ on THINGS-MEG, establishing SOTA
 across all settings. Further evaluation with pretrained EEG foundation models and other brain encoders shows that strong results are not limited to the a specific EEG encoder. Our EEG-to-image reconstruction pipeline also achieves SOTA on THINGS-EEG2 across all
seven reported metrics.

\end{itemize}

\section{Related Work}
\label{sec:related}

\paragraph{EEG and MEG visual decoding.}
Visual decoding aims to recover information about viewed images
from recorded brain activity.
EEG and MEG capture
visually evoked responses with millisecond-level temporal resolution. 
Two common tasks are image retrieval, which identifies the viewed
stimulus from a set of candidate images, and image reconstruction,
which generates an image from the recorded brain response.
Recent approaches address these tasks by aligning brain
representations with features from pretrained visual encoders. 
Many existing methods use global representations
from pretrained visual encoders, typically a final-layer CLS token or
a pooled feature vector \citep{song2024nice,li2024atm,zheng2026brainhive}.
NICE uses contrastive EEG--image alignment for zero-shot object recognition,
comparing each EEG embedding with one averaged image template for each unseen
class \citep{song2024nice}. ATM uses aligned EEG and CLIP embeddings for image
retrieval and conditions a diffusion model on the EEG embedding for image
reconstruction \citep{li2024atm}. Reconstruction methods use several forms of
visual supervision. CognitionCapturer learns separate EEG embeddings aligned
with image, text, and depth embeddings \citep{lan2025cognitioncapturer}.
NEED transfers a video-reconstruction model to unseen subjects and to
static-image reconstruction without retraining \citep{huang2025need}.
AVDE instead predicts multiscale visual tokens autoregressively from EEG
\citep{chen2026avde}. For cross-subject
retrieval, SATTC uses statistics from the unlabeled test batch to recalibrate
the EEG--image similarity matrix \citep{huang2026sattc}. For MEG,
\citet{csaky2023interpretable} study many-class image decoding, and
\citet{benchetrit2024brain} align MEG with pretrained visual embeddings for
image retrieval and reconstruction.

\paragraph{Constructing the visual embedding.}
Several methods construct the visual embedding from transformed images or
multiple pretrained representations. UBP chooses the blur applied to each
image from the estimated uncertainty of its paired EEG response, while Blur
Perception extracts CLIP features at multiple blur levels and learns to select
and adapt them \citep{wu2025ubp,liu2026blur}. NeuroBridge averages CLIP
features from several augmented views of each image
\citep{zhang2026neurobridge}. ViEEG constructs three CLIP embeddings from a
binary object mask, a foreground-masked image, and the full scene, to align
 with three EEG streams \citep{wang2025vieeg}. HyFI interpolates two CLIP
embeddings produced from foveally and Gaussian-blurred images in hyperbolic
space \citep{jo2026hyfi}. BrainHIVE combines pooled embeddings from
multiple CLIP encoders with a flattened VAE representation
\citep{zheng2026brainhive}. Our comparison focuses on a
single frozen layer and varies how its patch tokens are transformed before
pooling.

\paragraph{Intermediate visual representations.}
Recent work examines visual-encoder layer selection and patch-token
aggregation for EEG-to-image alignment.
\citep{du2026shallow} compares pooling methods over intermediate patch tokens.
\citep{tang2026aligning} averages patch tokens and combines embeddings
from two intermediate layers.
\citep{samga2026} combines intermediate-layer embeddings using
subject-specific weights.
Building on these studies, we focus on adapting the complete
patch-token set in to a visual embedding for alignment.

\section{Method}
\label{sec:method}

\subsection{Task formulation}

Let $\mathcal{D}=\{(b_i,I_i)\}_{i=1}^{n}$ contain paired visually evoked
brain responses and stimulus images, where
$b_i\in\mathbb{R}^{C_b\times T}$ and
$I_i\in\mathbb{R}^{H\times W\times3}$. Here $C_b$ is the number of brain-signal
channels, $T$ is the number of time samples, and $H$ and $W$ are the image
height and width. The brain encoder and visual adapter
produce $d$-dimensional vectors $h_{b,i}$ and $h_{v,i}$. We call
these the M/EEG embedding and the visual embedding.
For image retrieval, we rank the set of test images by cosine similarity to the brain embedding. 
Appendix~\ref{app:protocol} specifies the datasets and retrieval protocols.

\subsection{Learning a visual embedding from intermediate patch tokens}

\paragraph{Visual-encoder layer and visual adapter.}
We define a visual embedding by selecting an intermediate layer $\ell$ from a
frozen visual encoder $F$ and applying a trainable visual adapter
$G_\theta$ to its output. Thus
$h_{v,i}=G_\theta(F_\ell(I_i))$ is the visual embedding used for alignment.
Throughout the paper, \emph{visual adapter} refers to the complete
trainable image-side mapping $G_\theta$, whereas \emph{visual embedding}
refers to the normalized vector $z_{v,i}$ formed from that adapter's output. 
At a selected intermediate layer, a VIT based frozen visual encoder returns a CLS token
and a grid of patch tokens,
\begin{equation}
    F_{\ell}(I)=[c;X]\in\mathbb{R}^{(N+1)\times D},
\end{equation}
where $c$ is the CLS token and
$X=(x_1,\ldots,x_N)$ contains $N$ patch tokens arranged
on an $H_{\ell}\times W_{\ell}$ grid. The visual encoder is kept frozen. 
Appendix~\ref{app:protocol} gives the  full configuration for M/EEG-to-Image retrieval task .
\begin{figure}[!htbp]
\centering
\includegraphics[width=\linewidth,trim=0 0 0 0,clip]
{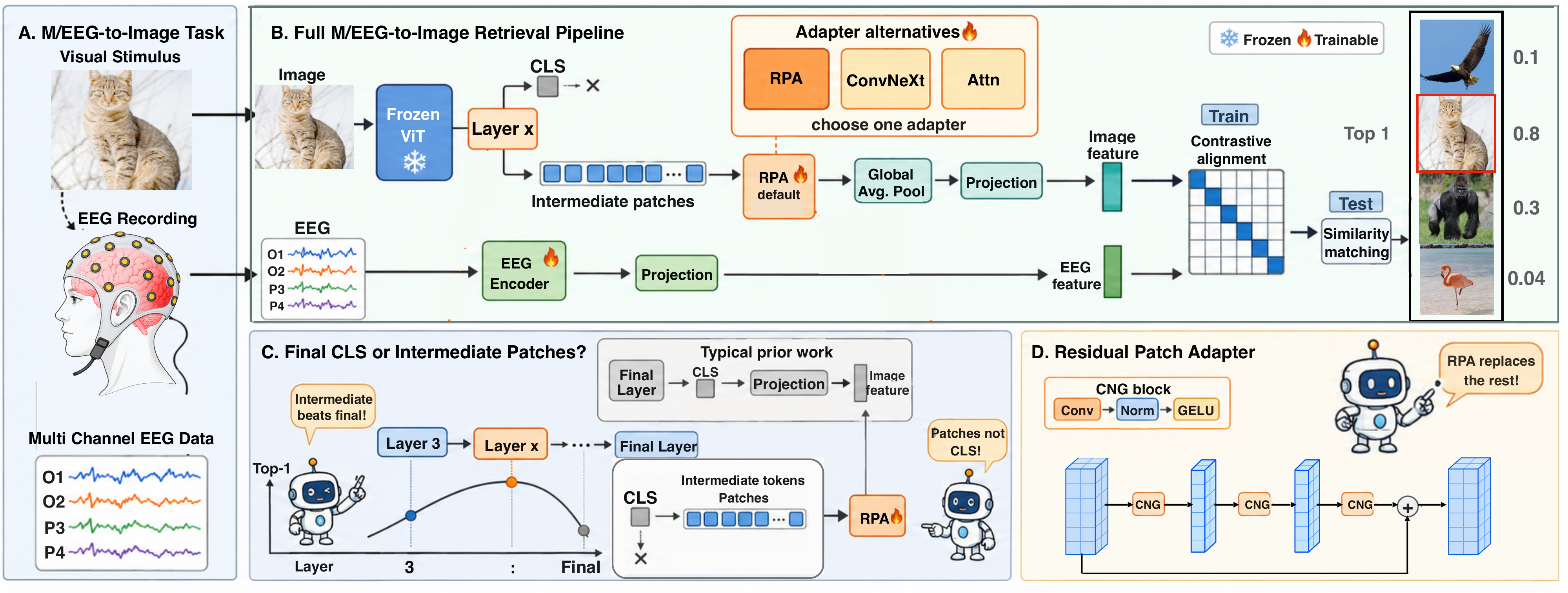}
\caption{Overview of the EEG-to-image retrieval framework. \textbf{(A)} EEG is
recorded while a participant views an image. \textbf{(B)} A frozen visual
encoder and a trainable visual adapter produce the visual embedding, which is
aligned with the brain embedding for retrieval. \textbf{(C)} The schematic
contrasts a final-layer CLS route with an intermediate-patch route; the
controlled comparison in Table~\ref{tab:central_ladder} uses CLS and patch
tokens from the same intermediate layer. \textbf{(D)} The residual adapter
branch transforms the patch-token grid with convolution--normalization--GELU
layers. Pooling and projection, shown separately in panel B, complete the
visual adapter.}
\label{fig:target_factorization}
\end{figure}

\paragraph{Visual adapters.}
The visual adapter $G_\theta$ maps the selected visual encoder
features to a learned representation for brain--image alignment.
In our approach, the adapter transforms the complete patch-token
set before pooling, allowing the visual representation to be
learned from individual patch tokens. A learned linear projection $q$ maps the aggregated
features to a $d$-dimensional vector $h_v$, which is normalized
to obtain the visual embedding $z_v$.

We implement this mapping with the \emph{Residual Patch Adapter}
(RPA). We discard the CLS token and reshape the patch tokens
into a $H_{\ell}\times W_{\ell}$ grid. RPA applies a residual convolutional branch
$A_\theta$ to this grid and adds its output to the original
patch-token features before pooling and projection.
Let $B_1$, $B_2$, and $B_3$ denote three blocks, each applying a
convolution followed by GroupNorm and GELU. Their kernel sizes are
$1\times1$, $3\times3$, and $1\times1$, respectively, with channel
dimensions $D\rightarrow d_a\rightarrow d_a\rightarrow D$,
where $d_a$ is the bottleneck width.
The residual branch and complete visual adapter are
\begin{equation}
\begin{aligned}
A_\theta(X) &= B_3\bigl(B_2(B_1(X))\bigr),\\
G_\theta(X) &= q\bigl(\operatorname{GAP}(X+A_\theta(X))\bigr).
\end{aligned}
\end{equation}
Here $\operatorname{GAP}$ denotes global average pooling over the
patch grid, and $q$ is the learned linear projection.
Appendix~\ref{app:operator_results} provides a set of  alternative adapters.

\subsection{Brain--image alignment}

\paragraph{Brain encoder.}
We adopt EEGProjection brain encoder to map each M/EEG response
to a $d$-dimensional representation for alignment with the visual embedding.
The brain encoder is trained jointly with the visual adapter
through contrastive learning, while the pretrained visual encoder
remains frozen. The implementation details are provided in the appendix~\ref{app:protocol}. 

\paragraph{MEG channel projection.}
THINGS-MEG records $C$ channels, making the flattened input to the brain
encoder large. For an MEG trial $b\in\mathbb{R}^{C\times T}$, we learn a
bias-free matrix $M\in\mathbb{R}^{K\times C}$ and compute
\begin{equation}
\widetilde b=Mb.
\end{equation}
This produces $\widetilde b\in\mathbb{R}^{K\times T}$. We pass
$\widetilde b$ to the brain encoder; EEG inputs do not use this operation.
Appendix~\ref{app:meg} reports the effect of the projection output dimension.

\paragraph{Contrastive learning objective.}
We use a CLIP-style symmetric InfoNCE objective
\citep{radford2021clip} to align the brain and visual embeddings. For a
minibatch of $B$ paired normalized embeddings, the scaled cosine-similarity
logits are
\begin{equation}
s_{ij}=\exp(\gamma)h_{b,i}^{\top}h_{v,i},
\end{equation}
where $\gamma$ is a trainable logit scale. The alignment objective is
\begin{equation}
\mathcal{L}_{\mathrm{NCE}}
=-\frac{1}{2B}\sum_{i=1}^{B}
\left[
\log\frac{\exp(s_{ii})}{\sum_{j=1}^{B}\exp(s_{ij})}
+
\log\frac{\exp(s_{ii})}{\sum_{j=1}^{B}\exp(s_{ji})}
\right].
\end{equation}
The two terms perform brain-to-image and image-to-brain matching,
respectively. In each direction, the paired sample is the positive and the
remaining batch elements serve as negatives. We jointly optimize the brain
encoder, visual adapter and for MEG, the learned sensor
projection. The
supplement reports the detailed settings.

\subsection{EEG-to-image reconstruction}
\label{sec:generation_overview}

Beyond retrieval, we evaluate EEG-to-image reconstruction with a separately
trained dual-stream generation pipeline. Its brain encoder is aligned with the
RPA visual embedding and a pooled CLIP-H embedding; two conditional priors then
predict semantic and structural conditions for the image generator.
Appendix~\ref{app:generation} describes
the pipeline, compares alignment objectives, and reports quantitative and
qualitative results.

\section{Experiments}
\label{sec:experiments}
\label{sec:operator_study}

We evaluate 200-way image retrieval on THINGS-EEG2 and THINGS-MEG under
within-subject and leave-one-subject-out (LOSO) protocols. Unless stated
otherwise, each test input averages 80 repeated EEG responses or 12 repeated
MEG responses to the same image. EEG training averages four responses per
image; MEG training uses single presentations. Appendix~\ref{app:protocol}
describes the data, training configurations, and evaluation procedures.
In this section, we first provide our quantitative result for retrievals and reconstructions. Then We  examine visual-encoder depth, and compare adapters at a fixed image encoder layer and test patch-token ability for EEG decoding.We next provide results for reconstruction and visualization .
\subsection{Quantitative Evaluation}
\label{sec:quantitative_evaluation}
\label{sec:brain_encoder_generality}
\label{sec:prior_comparison}
\label{sec:end_to_end_retrieval}

We evaluate two tasks: M/EEG-to-image retrieval and EEG-to-image
reconstruction. For retrieval, we report 200-way zero-shot Top-1 and Top-5
accuracy on THINGS-EEG2 and THINGS-MEG under within-subject and
leave-one-subject-out (LOSO) protocols. For reconstruction, we evaluate
low-level similarity using PixCorr and SSIM, and feature-level agreement
using AlexNet-2/5, Inception, CLIP, and SwAV. Higher values are better for
all metrics except SwAV. Results are averaged across participants.

Our method achieves the highest Top-1 and Top-5 accuracies across all
four retrieval settings in Table~\ref{tab:headline_benchmark}.
On THINGS-EEG2, within-subject Top-1/Top-5 accuracy reaches
$95.4\%/99.8\%$, compared with $84.6\%/98.2\%$ for HCF, while
LOSO accuracy reaches $35.5\%/67.4\%$, compared with
$23.4\%/54.9\%$. On THINGS-MEG, our method achieves
$65.2\%/85.7\%$ within subject and $6.7\%/19.2\%$ under LOSO,
outperforming the reported baselines in both settings.
These improvements extend across EEG and MEG and across both evaluations. Subject-level results and additional
baselines are provided in Appendix~\ref{app:subject_benchmarks}.

For EEG-to-image reconstruction, we evaluate the separately trained
EEG encoder aligned with both the RPA visual embedding and the pooled
CLIP-H embedding, using the pipeline described in
Appendix~\ref{sec:generation_method}.
Table~\ref{tab:generation_benchmark} reports results across 10 THINGS-EEG2 participants. Our system obtains the best
reported value on all seven metrics.
Together, these results show higher reconstruction similarity at both
the pixel and feature levels.

We further examine whether RPA can be paired with alternative brain
encoders, including pretrained EEG foundation models.
Keeping the frozen visual encoder and RPA architecture unchanged,
we replace the EEGProjection encoder with conventional EEG foundation models.
EEG-Conformer (without pretraining) and CBraMod (pretrained) achieve
within-subject/LOSO Top-1 accuracies of $94.90\%/34.45\%$ and
$94.55\%/35.63\%$, respectively, comparable to those of the
EEGProjection encoder.
Full results are reported in Appendix~\ref{app:brain_operating}.
The results show that strong retrieval
performance with RPA is not limited to the EEGProjection and
extends to a set of EEG foundation model.

\begin{table}[!htbp]
\centering
\caption{Average Top-1/Top-5 accuracy (\%) for 200-way zero-shot retrieval on
THINGS-EEG2 and THINGS-MEG. Values are participant averages;
``--'' indicates not reported.}
\label{tab:headline_benchmark}
\scriptsize
\setlength{\tabcolsep}{2.4pt}
\resizebox{\linewidth}{!}{%
\begin{tabular}{@{}lrrrrrrrr@{}}
\toprule
& \multicolumn{4}{c}{EEG} & \multicolumn{4}{c}{MEG} \\
\cmidrule(lr){2-5}\cmidrule(l){6-9}
& \multicolumn{2}{c}{Within} & \multicolumn{2}{c}{Cross}
& \multicolumn{2}{c}{Within} & \multicolumn{2}{c}{Cross} \\
Method & T1 & T5 & T1 & T5 & T1 & T5 & T1 & T5 \\
\midrule
NICE \citep{song2024nice}
 & 13.8 & 39.5 & 6.2 & 21.4 & 12.8 & 36.0 & -- & -- \\
ATM \citep{li2024atm}
 & 28.6 & 58.5 & 11.8 & 33.7 & -- & -- & -- & -- \\
UBP \citep{wu2025ubp}
 & 50.9 & 79.7 & 12.4 & 33.4 & 26.7 & 55.2 & 2.2 & 10.4 \\
BrainHIVE \citep{zheng2026brainhive}
 & 75.7 & 94.6 & 20.0 & 44.1 & 33.7 & 60.5 & 5.4 & 15.2 \\
Blur Perception \citep{liu2026blur}
 & 80.0 & 96.9 & 20.0 & 48.0 & 44.0 & 72.0 & 5.3 & 15.9 \\
Shallow Alignment \citep{du2026shallow}
 & 82.6 & 97.7 & 21.8 & 49.4 & 48.0 & 74.4 & 3.7 & 13.0 \\
HCF \citep{tang2026aligning}
 & 84.6 & 98.2 & 23.4 & 54.9 & -- & -- & -- & -- \\
\midrule
\textbf{Ours}
 & $\mathbf{95.4{\pm}0.1}$ & $\mathbf{99.8{\pm}0.0}$
 & $\mathbf{35.5{\pm}0.3}$ & $\mathbf{67.4{\pm}0.7}$
 & $\mathbf{65.2{\pm}1.0}$ & $\mathbf{85.7{\pm}0.8}$
 & $\mathbf{6.7{\pm}0.4}$ & $\mathbf{19.2{\pm}1.0}$ \\
\bottomrule
\end{tabular}}
\end{table}

\begin{table}[!htbp]
\centering
\caption{Average image-reconstruction performance on THINGS-EEG2.}
\label{tab:generation_benchmark}
\scriptsize
\setlength{\tabcolsep}{3pt}
\resizebox{\linewidth}{!}{%
\begin{tabular}{@{}lrrrrrrr@{}}
\toprule
& \multicolumn{2}{c}{Low-level} & \multicolumn{5}{c}{High-level} \\
\cmidrule(lr){2-3}\cmidrule(l){4-8}
Method & PixCorr $\uparrow$ & SSIM $\uparrow$ & AlexNet-2 $\uparrow$
& AlexNet-5 $\uparrow$ & Inception $\uparrow$ & CLIP $\uparrow$ & SwAV $\downarrow$ \\
\midrule
CognitionCapturer (All) \citep{lan2025cognitioncapturer}
& 0.150 & 0.347 & 0.754 & 0.623 & 0.669 & 0.715 & 0.590 \\
AVDE \citep{chen2026avde}
& 0.147 & 0.366 & 0.766 & 0.835 & 0.724 & 0.747 & 0.587 \\
BrainHIVE \citep{zheng2026brainhive}
& 0.195 & 0.336 & 0.843 & 0.905 & 0.756 & 0.808 & 0.554 \\
\midrule
\textbf{Ours}
& \textbf{0.225} & \textbf{0.454} & \textbf{0.865} & \textbf{0.909}
& \textbf{0.783} & \textbf{0.811} & \textbf{0.526} \\
\bottomrule
\end{tabular}}
\end{table}

\subsection{From Images to Intermediate Visual Representations}
\label{sec:depth_results}

\begin{figure}[!htbp]
\centering
\includegraphics[width=\linewidth]{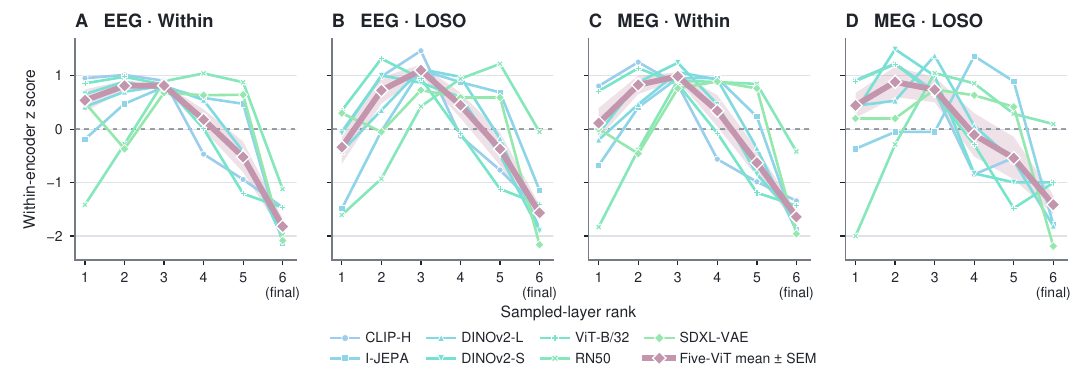}
\caption{Visual-encoder depth profiles for EEG and MEG under within-subject and
LOSO evaluation. The x-axis shows six uniformly spaced relative depths. Curves show mean 200-way Top-1 scores, averaged across three
independent runs and standardized within each encoder.
}
\label{fig:depth_uniform6}
\end{figure}

In Figure~\ref{fig:depth_uniform6}, we compare seven frozen visual encoders, OpenCLIP ViT-H/14, CLIP ViT-B/32 and
RN50 \citep{radford2021clip,ilharco2021openclip}, I-JEPA
\citep{assran2023ijepa}, DINOv2-S/L \citep{oquab2024dinov2}, and SDXL-VAE
\citep{podell2024sdxl}, at six relative depths under within-subject and LOSO
on M/EEG. Scores are
standardized within each encoder.  Although the best
depth varies across image encoders, the broadly inverted-U-shaped depth shape appears in all image encoders. This trend aligns with the hierarchical progression of visual
representations across encoder layers, in which intermediate layers combine low- and high-level image information. In contrast, deeper layers become more specialized for
pretraining objectives, such asimage--text alignment, self-supervised
learning, or reconstruction.
Appendix~\ref{app:depth_backbones} reports absolute scores and procedures.

\subsection{From Patch Tokens to a Visual Embedding}
\label{sec:main_ladder}

We next examine whether adapting the complete patch-token set produces
a more effective visual embedding than using the CLS token or the
patch mean. We compare the CLS-token
and patch-mean baselines with pointwise RPA and RPA.
As shown in Table~\ref{tab:central_ladder}, the patch-mean 
achieves higher accuracy than the CLS-token , while
pointwise RPA provides a further substantial improvement.
These results show that adapting the complete patch-token set before
pooling yields higher retrieval accuracy than using the
CLS token or patch mean.

We further examine how RPA compares with alternative visual adapters.
Using the same frozen patch-token input, we compare local RPA with
attention pooling,  transformer-based adapter, ConvNeXt, and
depthwise-separable convolution.
As reported in Appendix, local RPA achieves the highest mean
accuracy in both within-subject and LOSO evaluation.
The alternative learned adapters have more trainable parameters
but lower accuracies, supporting RPA as a lightweight and effective
visual adapter for EEG-to-image alignment. The adapters evaluated in these ablations also form a plug-and-play
toolkit for frozen visual encoders.

\begin{table}[!htbp]
\centering
\caption{Controlled within-subject THINGS-EEG2 comparison at a fixed
visual-encoder output.}
\label{tab:central_ladder}
\scriptsize
\begin{tabular}{llr}
\toprule
Visual input & Visual adapter & Top-1 (\%) \\
\midrule
CLS token & LayerNorm + linear & $69.50\pm0.45$ \\
patch tokens & mean, LayerNorm + linear & $77.03\pm0.7$ \\
patch tokens & pointwise RPA & $92.98\pm0.2$ \\

patch tokens & local RPA
& $\mathbf{95.4\pm0.1}$ \\
\bottomrule
\end{tabular}
\end{table}

\subsection{Patch-Token Interventions and Visual-Feature Probes}
\label{sec:analysis}

To understand why RPA benefits from the complete patch-token set,
we first examine whether simply averaging more frozen tokens improves
their predictability from EEG. The plateau in this linear probe
motivates a further question: does learned adaptation still benefit
from retaining individual tokens when averaging more tokens provides
little improvement? We therefore use pooling and masking to test
whether merging or replacing tokens limits RPA, while permutation
examines the additional contribution of spatial arrangement.
Finally, we probe six image attributes to help interpret these
token-level findings in terms of the visual information predictable
from EEG.

\begin{figure}[!htbp]
\centering
\includegraphics[width=\linewidth]{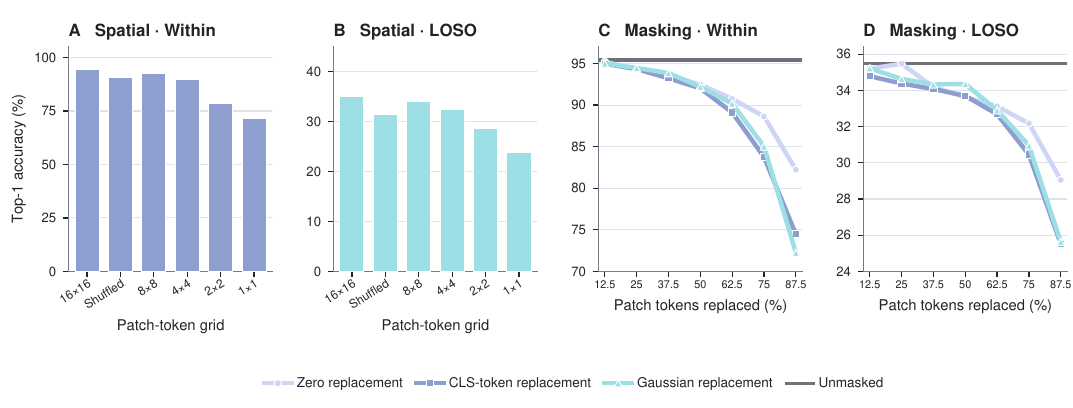}
\caption{\textbf{RPA diagnostics.} \textbf{(A--B)} Top-1 accuracy for native,
shuffled, and spatially averaged patch-token grids. \textbf{(C--D)} Top-1
accuracy under increasing zero, CLS-token, or Gaussian replacement. Solid gray
lines mark unmasked performance.}
\label{fig:rpa_diagnostics}
\end{figure}

\paragraph{Averaging frozen patch tokens.}
We next examine whether averaging more patch tokens produces visual
representations that are more predictable from EEG.
At a fixed visual-encoder layer, we average patch-token subsets of
different sizes. For each subset size, we fit a ridge
regression model to predict the averaged features from EEG and use
the predictions to retrieve the paired images.
This experiment uses frozen visual features without RPA.
As shown in Figure~\ref{fig:representation_views}\textbf{C},
averaging larger subsets yields higher Top-1 accuracy than using
a single patch token. These results suggest that aggregating
multiple patch tokens better captures visual information predictable
from EEG than using a single token. Although accuracy in this linear probe plateaus as more patch tokens
are averaged, our pooling and masking ablations in the follow section show
that RPA benefits from retaining and adapting the complete patch-token
set before pooling.
Full experimental details are provided in
Appendix~\ref{sec:app_frozen_probes}.

\paragraph{Patch-token pooling and spatial arrangement.}
\label{sec:topology}
At a fixed visual-encoder layer, we apply local average pooling to the
original $16\times16$ patch-token grid before training RPA. We use pooling
blocks of $2\times2$,..., $16\times16$, with the
last condition corresponding to global mean pooling. Larger pooling
blocks merge more patch tokens, reducing the number of distinct token
representations available to RPA when forming the visual embedding.
As shown in Figure~\ref{fig:rpa_diagnostics}\textbf{A--B}, Top-1 accuracy
decreases as the pooling block size increases in both within-subject and
LOSO evaluation. These results indicate retaining the full patch-token
set for adaptation provide enougn information to form visual embedding.

To examine the contribution of spatial arrangement, we also permute the
original patch tokens on the grid and retrain RPA using the permuted
inputs during both training and evaluation. This preserves every token's
values while disrupting the original spatial arrangement presented to
the adapter. Within-subject Top-1 accuracy decreases from $95.40\%$ to
$90.70\%$, indicating that preserving the original arrangement provides
an additional benefit for EEG-to-image retrieval.

\paragraph{Masking patch tokens.}
To complement the pooling experiment, we examine how retrieval accuracy
changes when RPA has access to fewer original patch tokens.
At the same fixed visual-encoder layer, we independently mask each patch
token with probability $p$, replacing it with the image's CLS token,
Gaussian noise, or a zero vector.
Unlike pooling, this operation preserves the patch-token grid while
replacing selected token values.
We retrain RPA for each condition, using the same masking rate and
replacement strategy during training and evaluation.

As shown in Figure~\ref{fig:rpa_diagnostics}\textbf{C--D}, none of the
masking conditions outperforms the unmasked baseline, and Top-1 accuracy
generally decreases as the masking rate increases in both within-subject
and LOSO evaluation.
At high masking rates, zero replacement outperforms CLS-token
replacement, indicating that filling masked positions with the CLS
token provides no advantage over leaving them zero.
Together with the pooling experiment,  these experiments show that preserving the complete set of patch-token  for RPA provide all information needed for EEG decoding.

\paragraph{Image attributes predicted from EEG.}
We next examine which image attributes can be predicted from EEG.
We represent each stimulus image using features describing six image
attributes: (1) object-category semantics, (2) color, (3) texture,
(4) objectness and two-dimensional layout, (5) oriented edges,
and (6) object silhouette.
For each attribute, we fit a separate ridge regression model to
predict the corresponding image features from EEG, then use the
predicted features to retrieve the paired images.
As reported in Appendix~\ref{sec:app_named_feature_probes},
object-category semantics, color, and texture yield the highest
Top-1 accuracies.
These results indicate that image attributes predictable from EEG
extend beyond object-category semantics to lower-level visual
properties, particularly color and texture.
These findings may help explain the strong EEG-to-image decoding
performance achieved using the complete set of patch tokens from
an intermediate layer.

\subsection{EEG-to-Image Reconstruction and Alignment Diagnostics}
\label{sec:additional_diagnostics}

\paragraph{EEG-to-image reconstruction.}
\label{sec:reconstruction_main}

\begin{figure}[!htbp]
\centering
\includegraphics[width=\linewidth]{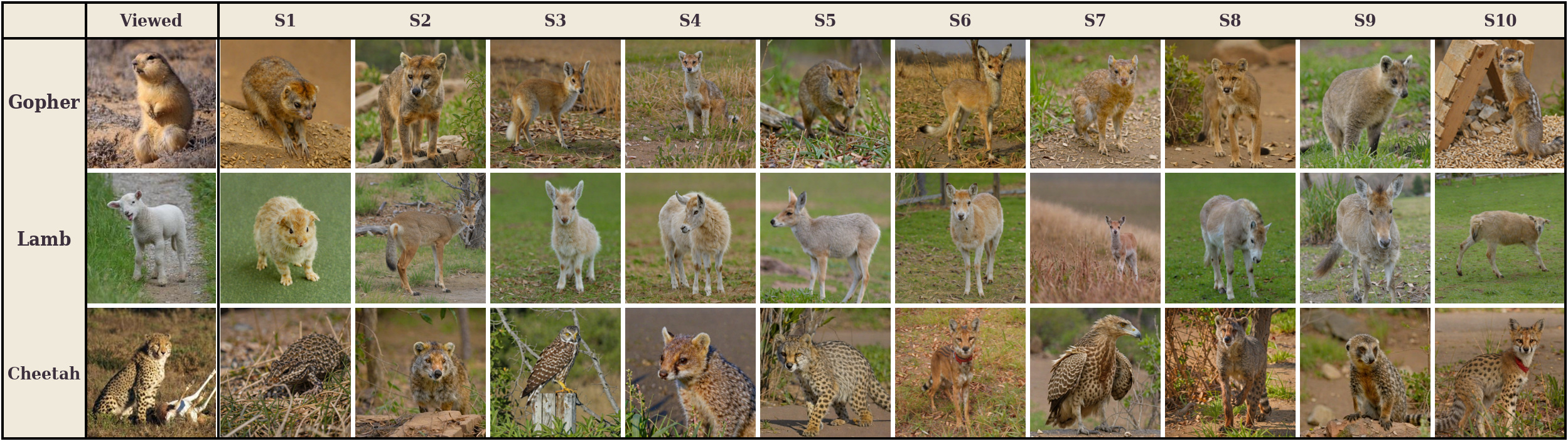}
\caption{\textbf{Qualitative EEG-to-image reconstructions.} Each row shows one
animal concept: the viewed image followed by
reconstructions from Subjects~1--10.}
\label{fig:generation_examples}
\end{figure}

Figure~\ref{fig:generation_examples} shows the viewed image and all 10
participant reconstructions for three  animal concepts.
Additional examples  are reported in
the Appendix~\ref{sec:app_generation_qualitative}.

\begin{figure}[!htbp]
\centering
\includegraphics[width=\linewidth]{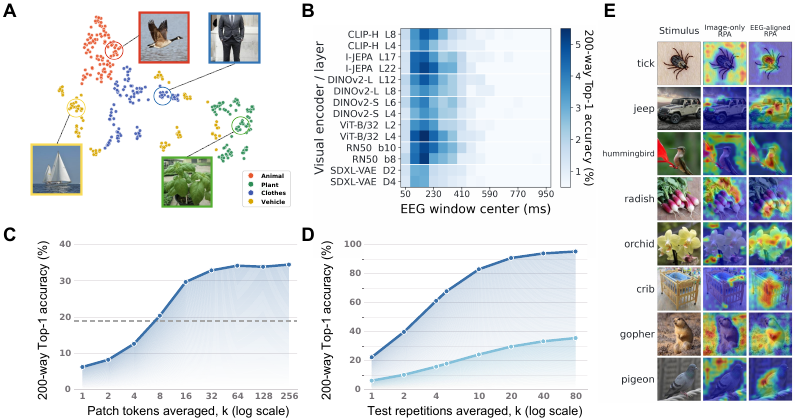}
\caption{\textbf{Alignment, timing, averaging, and attribution diagnostics.}
\textbf{(A)} UMAP projection of EEG embeddings; colors indicate categories, and insets show example
stimuli. \textbf{(B)} Given layers of visual encoder and 100-ms EEG
windows advanced in 60-ms steps, we using a ridge probe to obtain
a decoding accuracy. \textbf{(C)}
Retrieval with ridge-predicted means of frozen patch-token subsets;
the dashed line marks the CLS-token result. \textbf{(D)} Top-1 accuracy versus the number of EEG test
trials averaged per stimulus;
dark and light blue denote within-subject and LOSO evaluation. \textbf{(E)}
Grad-CAM maps; columns show the stimulus, image-only RPA, and RPA.}
\label{fig:representation_views}
\end{figure}

\paragraph{Visualizing the aligned EEG embedding space.}
\label{sec:umap-vis}
We extract EEG embeddings for all test images from each
subject's within-subject model. For visualization, we align
them across participants using orthogonal Procrustes transformations
fitted to the corresponding visual embeddings. 
We align EEG embeddings from four categories images and
project them to two dimensions using UMAP \citep{mcinnes2018umap}.
Each point in Figure~\ref{fig:representation_views}\textbf{A}
represents one subject--image pair. The details of this experiment goes into the appendix~\ref{sec:app_umap_vis}.
The plot shows category-level organization in this selected subset.
Although the model is not explicitly trained for four-category
classification, the aligned EEG embeddings tend to cluster by
category in the visualization. This suggests that EEG embeddings learned through RPA capture category-level semantic structure. 

\paragraph{Visual-encoder depth and EEG latency.}
\label{sec:depth_time_main}
We next examine whether visual representations from deeper encoder
layers are best predicted from later EEG activity.
We use ridge regression probes to predict frozen visual features
from successive EEG windows.
As shown in Figure~\ref{fig:representation_views}\textbf{B},
all selected layers reach their highest decoding accuracy in
windows centered between 110 and 230\,ms, with most peaking
at 170\,ms, which temporally overlaps with the
N170, an occipitotemporal component associated with early face
processing \citep{bentin1996electrophysiological}.
Accuracy then declines and approaches the chance level
after 400\,ms.
These results indicate that visual representations are most predictable from EEG within early intervals.
Full experimental details are provided in
Appendix~\ref{sec:app_depth_time}.

\paragraph{Grad-CAM visualization.}
\label{sec:gradcam}
We next examine which image regions RPA highlights under EEG
alignment and image-only training using Grad-CAM
\citep{selvaraju2017gradcam}.
For image-only RPA, the maps are computed from similarity to the
final CLIP image embedding.
Figure~\ref{fig:representation_views}\textbf{E} compares the
normalized Grad-CAM maps for eight selected test images.
In these examples, EEG-aligned RPA produces maps concentrated on
foreground objects, whereas image-only RPA shows less foreground
focus. This contrast suggests that EEG alignment helps RPA
emphasize foreground-related visual information.
The visualization procedure are
provided in Appendix~\ref{sec:app_gradcam}.

\paragraph{Test-repetition averaging}
\label{sec:repetitions}
Recall that for test data, each data clip is the mean over 80 test trial given a fixed image stimulus. In Figure~\ref{fig:representation_views}\textbf{D} we do a experiemnt for a test count
$k$, repetitions are partitioned into disjoint groups
of size $k$, each group is averaged, and group accuracies are pooled .
Single-test-trial accuracy exceeds chance, and accuracy increases as more
responses are averaged.

\section{Conclusion}
\label{sec:conclusion}

This study proposed the Residual Patch Adapter (RPA), a novel lightweight visual adapter that moves beyond global embeddings to learn visual representations from the complete patch-token set. Ablation studies involving intermediate layer analysis, patch-token pooling and masking demonstrate the effectiveness of RPA, which achieves state-of-the-art performance in both within-subject and leave-one-subject-out M/EEG-to-image retrieval. The contributions of semantics, color, and texture to EEG decoding further support RPA’s use of diverse visual features. We further develop a suite of adapters that provides a plug-and-play interface for various visual encoders. Together, our work offers an effective and flexible approach to representation learning for brain-to-image decoding, with significant implications for brain--computer interface applications and future research on brain--image alignment.

\subsection*{AI use statement}

We used large language models (LLMs) to assist with manuscript drafting and
revision, including grammar checking, language editing, and feedback on clarity
and paragraph organization. The authors remain responsible for evaluating
AI-assisted suggestions and verifying the text, references, and numerical
claims against original sources and experimental records. The authors take full
responsibility for the final manuscript, research decisions, and reported results.

\subsection*{Ethics statement}

This study analyzes previously released THINGS-EEG2 and THINGS-MEG data and
introduces no new data collection or intervention with human participants. We
follow the datasets' terms of use and report aggregate and participant-level
retrieval results, together with image reconstructions. The evaluation
concerns responses to controlled visual stimuli, and the strongest results
use averages over repeated presentations. These conditions limit conclusions
about unrestricted or real-time brain decoding.

\FloatBarrier
\bibliographystyle{plainnat}
\bibliography{references}

\ifwithsupplement
\clearpage
\appendix
\section{Datasets, Training, and Evaluation Protocols}
\label{app:protocol}

This appendix describes the datasets, training configurations, and evaluation
procedures used in Section~\ref{sec:method} and the experiments. The matched
adapter comparison, benchmark system, and supplementary sweeps are reported
separately; their reference accuracies should be interpreted within the
configurations and replication schemes stated for each experiment.

\subsection{Datasets and preprocessing}

\paragraph{THINGS-EEG2.}
\citep{gifford2022things}
The primary dataset contains recordings from 10 participants. The distributed
data are sampled at 250\,Hz, preprocessed, and whitened. The training split has
16,540 images from 1,654 concepts; each image has four repetitions. The test
split has 200 images from 200 unseen concepts; each image has 80 repetitions.
There is no train--test concept overlap. Unless an ablation states otherwise,
the four training repetitions and 80 test repetitions are averaged before the
model receives them. Each standard test input is therefore the mean of 80
responses to one image.
The default EEG input comprises 17 anatomically specified posterior channels:
P7, P5, P3, P1, Pz, P2, P4, P6, P8, PO7, PO3, POz, PO4, PO8, O1, Oz, and O2.
This set was selected by scalp location rather than retrieval performance.

\paragraph{THINGS-MEG.}
\citep{hebart2023things}
The MEG experiments use four participants, 271 sensors, 19,848
single-presentation training images, and 200 test images with 12 repetitions.
Test repetitions are averaged. The learned sensor projection is fit only from the
training split: within subject it is fit on that participant's training data;
under LOSO it is fit on the other participants. Both
benchmark MEG settings use projection output dimension $K=64$.

\subsection{Within-subject and LOSO training configurations}

Table~\ref{tab:app_protocol_recipes} gives the within-subject and LOSO
training configurations. They differ in the data split, input duration, and
optimization settings. The difference in their accuracies therefore reflects
these configuration changes as well as the subject-generalization task.

\begin{table}[!htbp]
\centering
\caption{Within-subject and LOSO EEG training configurations.}
\label{tab:app_protocol_recipes}
\small
\begin{tabular}{lll}
\toprule
Setting & Within subject & LOSO \\
\midrule
Training responses & 16,540 from one subject & 148,860 from nine subjects \\
Test subject & same subject, unseen concepts & held-out subject, unseen concepts \\
Default window & 0--700\,ms & 0--1,000\,ms \\
Epochs & 25 & 15 \\
Peak learning rate & $5\times10^{-4}$ & $3\times10^{-4}$ \\
Projection dropout & 0.5 & 0.7 \\
Batch size & 512 & 512 \\
Validation fraction & 3\% & 3\% \\
Checkpoint rule & best validation retrieval & best validation retrieval \\
\bottomrule
\end{tabular}
\end{table}

\subsection{Model and optimization details}

We adopt the EEGProjection encoder from UBP \citep{wu2025ubp}
to map each M/EEG response to a $d$-dimensional representation
for alignment with the visual embedding.

\paragraph{Visual input and adapter configuration.}
The default visual input is the cached OpenCLIP ViT-H/14
layer-11 output, containing one CLS token and a $16\times16$
grid of patch tokens with feature dimension $D=1{,}280$.
These tokens have already passed through transformer self-attention
and can encode information beyond their corresponding image patches.
RPA uses a bottleneck width of $d_a=256$, 32 groups in each
GroupNorm layer, and an output dimension of $d=1{,}024$.

The pointwise RPA control replaces the middle $3\times3$
convolution with a $1\times1$ convolution, giving kernel sizes
$(1,1,1)$. Although this removes explicit local mixing,
GroupNorm still couples patch positions by aggregating statistics
across them.

\paragraph{Optimization.}
We use AdamW with $\beta=(0.9,0.995)$, weight decay $0.05$,
gradient-norm clipping at $1.0$, and cosine learning-rate decay.
The logit scale is initialized as $\gamma_0=\log(1/0.07)$,
corresponding to an initial temperature of $0.07$.

\subsection{Retrieval evaluation and model selection}

For each brain query, evaluation ranks the 200 held-out stimulus images by
cosine similarity.
Within-subject evaluation trains a separate model for each subject and
evaluates it on that subject's unseen test concepts. Leave-one-subject-out
(LOSO) evaluation trains on the remaining participants and evaluates the held-out
participant.
For each fixed configuration, we evaluate the test set at the
validation-selected checkpoint. 
For benchmark results, accuracy is first averaged across participants within
each training run; $\pm$ denotes the SD across the three participant-averaged
run means. 

\subsection{Training and Retrieval Algorithm}
\label{sec:rpa_algorithm}

Algorithm~\ref{alg:rpa_training_retrieval} summarizes RPA training and
M/EEG-to-image retrieval. We first extract and cache the complete
patch-token grid from a fixed layer of the frozen visual encoder,
discarding the CLS token. RPA transforms this grid before global
average pooling and projection. We jointly train RPA and the brain
encoder with symmetric contrastive learning, together with the logit
scale and, for MEG, the sensor projection. After restoring the
checkpoint with the highest validation retrieval accuracy, we encode
each candidate image and rank the candidates by cosine similarity
to the query's brain embedding.

Let $f_\phi$ denote the brain encoder and
$\nu(a)=a/\|a\|_2$.
The trainable parameters are $\Phi=\{\theta,\phi,\gamma\}$ for EEG,
with $M$ added for MEG; $\theta$ includes the residual branch and
projection $q$.
$\operatorname{PatchGrid}$ discards the CLS token, retains all patch
tokens, and reshapes them to $D\times H_\ell\times W_\ell$.
Each $\mathrm{CNG}^{a\leftarrow b}_{k\times k}$ maps $b$ channels
to $a$ using a convolution followed by GroupNorm(32) and GELU;
$d_a$ is the bottleneck width.
GAP denotes global average pooling, and $\operatorname{CE}$
denotes mean row-wise cross-entropy.
The procedure uses the preprocessing, regularization, and
optimization settings of the corresponding within-subject
or LOSO protocol.

\begin{breakablealgorithm}
\caption{RPA training and M/EEG-to-image retrieval}
\label{alg:rpa_training_retrieval}
\footnotesize
\begin{algorithmic}[1]

\Require Training and validation pairs
         $\mathcal D_{\mathrm{tr}},\mathcal D_{\mathrm{val}}$;
         frozen visual encoder $F$ and fixed layer $\ell$
\Require Epoch count $E$;
         candidate images $\{I_j^{\mathrm{cand}}\}_{j=1}^{J}$;
         query response $b^\star$
\Ensure Trained parameters $\Phi$ and ranked candidate-image
        indices $\pi$

\Function{RPAEmbed}{$X;\theta$}
    \State $U_1\gets
        \mathrm{CNG}^{d_a\leftarrow D}_{1\times1}(X)$
    \State $U_2\gets
        \mathrm{CNG}^{d_a\leftarrow d_a}_{3\times3}(U_1)$
    \State $U_3\gets
        \mathrm{CNG}^{D\leftarrow d_a}_{1\times1}(U_2)$
    \State \Return
        $\nu\!\left(q\!\left(
        \operatorname{GAP}(X+U_3)\right)\right)$
\EndFunction

\Needspace{3\baselineskip}
\Statex \textbf{Frozen patch-token extraction}
\State Cache
       $X(I)\gets\operatorname{PatchGrid}(F_\ell(I))$
       for all training and validation images

\Needspace{3\baselineskip}
\Statex \textbf{Joint training}
\State Initialize $\Phi$ with $\gamma=\log(1/0.07)$;
       keep $F$ frozen throughout

\For{$e=1,\ldots,E$}
    \State Set trainable modules to training mode

    \For{each minibatch
         $\{(b_i,I_i)\}_{i=1}^{B}$
         from $\mathcal D_{\mathrm{tr}}$}

        \State $\widetilde b_i\gets Mb_i$ for MEG,
               and $\widetilde b_i\gets b_i$ for EEG

        \State $z_{b,i}\gets\nu(f_\phi(\widetilde b_i))$,
               \quad
               $z_{v,i}\gets
               \Call{RPAEmbed}{X(I_i);\theta}$
               for each $i$

        \State $S_{ij}\gets
               \exp(\gamma)\,z_{b,i}^{\top}z_{v,j}$
               for $i,j=1,\ldots,B$

        \State $\mathbf y\gets(1,\ldots,B)$

        \State $\mathcal L\gets
               \tfrac12\!\left[
               \operatorname{CE}(S,\mathbf y)
               +\operatorname{CE}(S^{\top},\mathbf y)
               \right]$

        \State Update $\Phi$ with AdamW after clipping
               the gradient norm to $1.0$
    \EndFor

    \State Evaluate retrieval on $\mathcal D_{\mathrm{val}}$
           in evaluation mode; save the best checkpoint
\EndFor

\State Restore the best-validation checkpoint and set
       all modules to evaluation mode

\Needspace{3\baselineskip}
\Statex \textbf{Image retrieval}
\For{$j=1,\ldots,J$}
    \State $X_j^{\mathrm{cand}}\gets
           \operatorname{PatchGrid}
           (F_\ell(I_j^{\mathrm{cand}}))$

    \State $z_{v,j}^{\mathrm{cand}}\gets
           \Call{RPAEmbed}{X_j^{\mathrm{cand}};\theta}$
\EndFor

\State $\widetilde b^\star\gets Mb^\star$ for MEG,
       and $\widetilde b^\star\gets b^\star$ for EEG

\State $z_b^\star\gets\nu(f_\phi(\widetilde b^\star))$

\State $\pi\gets
       \operatorname{argsort}_{j}\!\left[
       (z_b^\star)^{\top}z_{v,j}^{\mathrm{cand}}
       \right]$
       in descending order

\State \Return $\Phi,\pi$

\end{algorithmic}
\end{breakablealgorithm}

\section{Subject-Level Benchmark Comparisons}
\label{app:subject_benchmarks}

\newcommand{\subjmsd}[2]{\shortstack{$#1$\\[-1pt]{\tiny$\pm#2$}}}
\newcommand{\subjmsdb}[2]{\shortstack{$\mathbf{#1}$\\[-1pt]{\tiny$\mathbf{\pm#2}$}}}

These tables report subject-level Top-1/Top-5 retrieval. Baselines are published
values; ours is mean$\,\pm\,$SD across three repeated training runs, ``--'' is
unreported.

\begin{table}[!htbp]
\centering
\caption{Subject-level within-subject retrieval on THINGS-EEG2.}
\label{tab:supp_eeg_intra_subjects}
\footnotesize
\setlength{\tabcolsep}{2.0pt}
\begin{tabular}{@{}l*{11}{r}@{}}
\toprule
Method & S1 & S2 & S3 & S4 & S5 & S6 & S7 & S8 & S9 & S10 & Mean \\
\midrule
\multicolumn{12}{@{}l}{\textit{Top-1 accuracy (\%)}} \\
NICE \citep{song2024nice} & 12.3 & 10.4 & 13.1 & 16.4 & 8.0 & 14.1 & 15.2 & 20.0 & 13.3 & 14.9 & 13.8 \\
UBP \citep{wu2025ubp} & 41.2 & 51.2 & 51.2 & 51.1 & 42.2 & 57.5 & 49.0 & 58.6 & 45.1 & 61.5 & 50.9 \\
ViEEG \citep{wang2025vieeg} & 34.1 & 38.4 & 40.6 & 50.1 & 28.9 & 44.3 & 38.6 & 54.0 & 37.3 & 42.8 & 40.9 \\
NeuroBridge \citep{zhang2026neurobridge} & 50.0 & 63.2 & 61.6 & 61.4 & 54.8 & 69.7 & 62.7 & 71.2 & 64.0 & 73.6 & 63.2 \\
BrainHIVE \citep{zheng2026brainhive} & 64.3 & 76.3 & 74.0 & 67.0 & 68.0 & 81.5 & 76.8 & 84.8 & 76.8 & 87.3 & 75.7 \\
Blur \citep{liu2026blur} & 81.9 & 81.7 & 78.3 & 76.9 & 71.4 & 84.3 & 78.2 & 84.5 & 78.7 & 84.1 & 80.0 \\
Shallow Align. \citep{du2026shallow} & 75.0 & 87.5 & 83.2 & 79.5 & 74.6 & 89.9 & 78.5 & 86.9 & 81.3 & 89.3 & 82.6 \\
HCF \citep{tang2026aligning} & 81.9 & 86.0 & 84.3 & 82.2 & 74.5 & 90.4 & 82.8 & 90.9 & 79.7 & 93.1 & 84.6 \\
\textbf{Ours} & \subjmsdb{96.8}{0.5} & \subjmsdb{96.3}{0.2} & \subjmsdb{97.2}{0.5} & \subjmsdb{91.5}{0.7} & \subjmsdb{92.5}{1.1} & \subjmsdb{95.7}{0.6} & \subjmsdb{96.2}{0.6} & \subjmsdb{97.7}{1.4} & \subjmsdb{93.7}{0.2} & \subjmsdb{96.8}{0.2} & \subjmsdb{95.4}{0.1} \\
\midrule
\multicolumn{12}{@{}l}{\textit{Top-5 accuracy (\%)}} \\
NICE& 36.6 & 33.9 & 39.0 & 47.0 & 26.9 & 40.6 & 42.1 & 49.9 & 37.1 & 41.9 & 39.5 \\
UBP & 70.5 & 80.9 & 82.0 & 76.9 & 72.8 & 83.5 & 79.9 & 85.8 & 76.2 & 88.2 & 79.7 \\
ViEEG & 71.3 & 67.9 & 74.7 & 80.8 & 61.5 & 76.5 & 75.2 & 82.5 & 74.9 & 79.8 & 74.5 \\
NeuroBridge & 77.6 & 90.6 & 91.1 & 90.0 & 85.0 & 92.9 & 88.8 & 95.1 & 91.0 & 97.1 & 89.9 \\
BrainHIVE & 88.8 & 95.3 & 95.0 & 91.8 & 91.5 & 96.3 & 96.8 & 98.5 & 95.8 & 99.3 & 94.6 \\
Blur & 96.8 & 96.8 & 96.0 & 97.2 & 93.7 & 98.7 & 96.7 & 98.1 & 96.8 & 98.1 & 96.9 \\
Shallow Align. & 94.3 & 98.9 & 98.2 & 96.1 & 96.4 & 99.3 & 97.3 & 99.4 & 97.8 & 99.1 & 97.7 \\
HCF & 98.3 & 98.9 & 97.7 & 98.8 & 94.1 & 98.9 & 97.2 & 99.4 & 98.4 & 99.8 & 98.2 \\
\textbf{Ours} & \subjmsdb{99.7}{0.2} & \subjmsdb{99.7}{0.2} & \subjmsdb{100.0}{0.0} & \subjmsdb{99.5}{0.4} & \subjmsdb{99.8}{0.2} & \subjmsdb{99.8}{0.2} & \subjmsdb{100.0}{0.0} & \subjmsdb{100.0}{0.0} & \subjmsdb{99.8}{0.2} & \subjmsdb{100.0}{0.0} & \subjmsdb{99.8}{0.0} \\
\bottomrule
\end{tabular}
\end{table}

\begin{table}[!htbp]
\centering
\caption{Subject-level LOSO retrieval on THINGS-EEG2.}
\label{tab:supp_eeg_loso_subjects}
\footnotesize
\setlength{\tabcolsep}{2.0pt}
\begin{tabular}{@{}l*{11}{r}@{}}
\toprule
Method & S1 & S2 & S3 & S4 & S5 & S6 & S7 & S8 & S9 & S10 & Mean \\
\midrule
\multicolumn{12}{@{}l}{\textit{Top-1 accuracy (\%)}} \\
NICE \citep{song2024nice} & 7.6 & 5.9 & 6.0 & 6.3 & 4.4 & 5.6 & 5.6 & 6.3 & 5.7 & 8.4 & 6.2 \\
UBP \citep{wu2025ubp} & 11.5 & 15.5 & 9.8 & 13.0 & 8.8 & 11.7 & 10.2 & 12.2 & 15.5 & 16.0 & 12.4 \\
ViEEG \citep{wang2025vieeg} & 22.7 & 24.7 & 19.0 & 25.5 & 19.8 & 20.7 & 20.9 & 20.8 & 23.8 & 31.2 & 22.9 \\
NeuroBridge \citep{zhang2026neurobridge} & 23.2 & 21.2 & 13.2 & 17.0 & 14.5 & 25.0 & 15.3 & 20.1 & 13.7 & 27.2 & 19.0 \\
BrainHIVE \citep{zheng2026brainhive} & -- & -- & -- & -- & -- & -- & -- & -- & -- & -- & 20.0 \\
Blur \citep{liu2026blur} & 27.3 & 32.0 & 11.1 & 18.6 & 16.9 & 16.1 & 16.0 & 18.3 & 13.6 & 30.1 & 20.0 \\
Shallow Align. \citep{du2026shallow} & 23.5 & 30.6 & 10.0 & 19.5 & 18.1 & 22.7 & 18.6 & 17.3 & 23.0 & 34.4 & 21.8 \\
HCF \citep{tang2026aligning} & 28.5 & 22.5 & 20.0 & 20.5 & 18.5 & \textbf{29.5} & 20.0 & 20.0 & 21.0 & 34.0 & 23.4 \\
\textbf{Ours} & \subjmsdb{49.3}{1.3} & \subjmsdb{42.2}{2.5} & \subjmsdb{24.2}{2.4} & \subjmsdb{31.5}{0.7} & \subjmsdb{34.2}{1.2} & \subjmsd{27.0}{1.5} & \subjmsdb{35.2}{0.9} & \subjmsdb{29.8}{1.2} & \subjmsdb{35.0}{0.4} & \subjmsdb{46.3}{1.5} & \subjmsdb{35.5}{0.3} \\
\midrule
\multicolumn{12}{@{}l}{\textit{Top-5 accuracy (\%)}} \\
NICE & 22.8 & 20.5 & 22.3 & 20.7 & 18.3 & 22.2 & 19.7 & 22.0 & 17.6 & 28.3 & 21.4 \\
UBP & 29.7 & 40.0 & 27.0 & 32.3 & 33.8 & 31.0 & 23.8 & 32.2 & 40.5 & 43.5 & 33.4 \\
ViEEG & 53.5 & 52.5 & 48.4 & 54.1 & 47.3 & 49.3 & 49.4 & 46.8 & 52.7 & 60.3 & 51.4 \\
NeuroBridge & 52.4 & 49.3 & 36.5 & 45.3 & 37.7 & 55.0 & 45.1 & 44.9 & 36.5 & 56.3 & 45.9 \\
BrainHIVE & -- & -- & -- & -- & -- & -- & -- & -- & -- & -- & 44.1 \\
Blur & 54.9 & 63.0 & 34.5 & 46.0 & 40.7 & 43.2 & 45.4 & 49.0 & 41.2 & 62.1 & 48.0 \\
Shallow Align. & 53.2 & 60.4 & 28.1 & 48.3 & 45.2 & 49.8 & 46.0 & 46.1 & 54.8 & 62.0 & 49.4 \\
HCF & 61.0 & 56.5 & 45.5 & 54.0 & 46.5 & \textbf{60.5} & 53.5 & 53.5 & 48.5 & 69.0 & 54.9 \\
\textbf{Ours} & \subjmsdb{79.5}{0.8} & \subjmsdb{73.0}{0.8} & \subjmsdb{50.7}{1.3} & \subjmsdb{63.0}{1.9} & \subjmsdb{69.7}{1.3} & \subjmsd{59.2}{1.3} & \subjmsdb{63.7}{0.5} & \subjmsdb{62.5}{0.7} & \subjmsdb{72.2}{1.5} & \subjmsdb{80.8}{1.7} & \subjmsdb{67.4}{0.7} \\
\bottomrule
\end{tabular}
\end{table}

\FloatBarrier
\subsection{MEG}

\begin{table}[!htbp]
\centering
\caption{Subject-level within-subject retrieval on THINGS-MEG.}
\label{tab:supp_meg_intra_subjects}
\small
\setlength{\tabcolsep}{7pt}
\resizebox{\linewidth}{!}{%
\begin{tabular}{@{}lrrrrr@{}}
\toprule
Method & S1 & S2 & S3 & S4 & Mean \\
\midrule
\multicolumn{6}{@{}l}{\textit{Top-1 accuracy (\%)}} \\
NICE \citep{song2024nice} & 9.6 & 18.5 & 14.2 & 9.0 & 12.8 \\
UBP \citep{wu2025ubp} & 15.0 & 46.0 & 27.3 & 18.5 & 26.7 \\
ViEEG \citep{wang2025vieeg} & 16.6 & 37.4 & 30.6 & 17.2 & 25.5 \\
NeuroBridge \citep{zhang2026neurobridge} & 16.5 & 53.7 & 40.4 & 18.1 & 32.2 \\
BrainHIVE \citep{zheng2026brainhive} & 14.0 & 63.8 & 41.0 & 17.0 & 33.9 \\
Blur \citep{liu2026blur} & 25.4 & 78.1 & 37.3 & 35.3 & 44.0 \\
Shallow Align. \citep{du2026shallow} & 25.5 & 81.9 & 56.0 & 28.6 & 48.0 \\
\textbf{Ours} & $\mathbf{34.7\pm1.7}$ & $\mathbf{96.2\pm0.6}$ & $\mathbf{82.2\pm0.8}$ & $\mathbf{47.7\pm2.2}$ & $\mathbf{65.2\pm1.0}$ \\
\midrule
\multicolumn{6}{@{}l}{\textit{Top-5 accuracy (\%)}} \\
NICE & 27.8 & 47.8 & 41.6 & 26.6 & 36.0 \\
UBP & 38.0 & 80.5 & 59.0 & 43.5 & 55.2 \\
ViEEG & 50.3 & 79.1 & 73.9 & 49.5 & 63.2 \\
NeuroBridge & 41.6 & 85.3 & 73.2 & 43.1 & 60.8 \\
BrainHIVE & 31.8 & 91.8 & 78.3 & 41.0 & 60.9 \\
Blur & -- & -- & -- & -- & 72.0 \\
Shallow Align. & 54.5 & 97.4 & 87.5 & 58.3 & 74.4 \\
\textbf{Ours} & $\mathbf{67.7\pm1.9}$ & $\mathbf{99.8\pm0.2}$ & $\mathbf{97.2\pm1.0}$ & $\mathbf{78.0\pm0.4}$ & $\mathbf{85.7\pm0.8}$ \\
\bottomrule
\end{tabular}
}
\end{table}

\begin{table}[!htbp]
\centering
\caption{Subject-level LOSO retrieval on THINGS-MEG.}
\label{tab:supp_meg_loso_subjects}
\small
\setlength{\tabcolsep}{7pt}
\resizebox{\linewidth}{!}{%
\begin{tabular}{@{}lrrrrr@{}}
\toprule
Method & S1 & S2 & S3 & S4 & Mean \\
\midrule
\multicolumn{6}{@{}l}{\textit{Top-1 accuracy (\%)}} \\
UBP \citep{wu2025ubp} & 2.0 & 1.5 & 2.7 & 2.5 & 2.2 \\
NeuroBridge \citep{zhang2026neurobridge} & \textbf{4.3} & 3.6 & 3.0 & 2.5 & 3.4 \\
BrainHIVE \citep{zheng2026brainhive} & -- & -- & -- & -- & 5.4 \\
Blur \citep{liu2026blur} & 2.9 & 7.7 & 5.8 & \textbf{4.7} & 5.3 \\
Shallow Align. \citep{du2026shallow} & 1.3 & 6.6 & 5.4 & 1.5 & 3.7 \\
\textbf{Ours} & $3.0\pm1.1$ & $\mathbf{11.0\pm1.4}$ & $\mathbf{9.0\pm0.8}$ & $3.7\pm0.2$ & $\mathbf{6.7\pm0.4}$ \\
\midrule
\multicolumn{6}{@{}l}{\textit{Top-5 accuracy (\%)}} \\
UBP & 5.7 & 17.2 & 10.5 & 8.0 & 10.4 \\
NeuroBridge & \textbf{13.1} & 15.6 & 11.2 & 11.3 & 12.8 \\
BrainHIVE & -- & -- & -- & -- & 15.2 \\
Blur & -- & -- & -- & -- & 15.9 \\
Shallow Align. & 6.9 & 18.5 & 18.5 & 7.9 & 13.0 \\
\textbf{Ours} & $10.5\pm1.1$ & $\mathbf{28.2\pm1.9}$ & $\mathbf{26.0\pm3.1}$ & $\mathbf{12.3\pm1.2}$ & $\mathbf{19.2\pm1.0}$ \\
\bottomrule
\end{tabular}
}
\end{table}

\FloatBarrier
\section{Visual Adapters at a Fixed Visual-Encoder Output}
\label{app:operator_results}

Table~\ref{tab:central_ladder} reports the matched adapter comparison.
This appendix gives parameter counts, alternative adapter definitions and
results, and the bottleneck-width and kernel-placement experiments.

\subsection{Visual adapter families}
\label{sec:app_adapter_families}

\paragraph{Shared adapter input and output.}
Every adapter in Table~\ref{tab:app_head_families} receives the same frozen
OpenCLIP ViT-H/14 layer-11 output. Writing
\begin{equation}
s=[c;t],\qquad
t\in\mathbb{R}^{B\times256\times1280},\qquad
g=\operatorname{grid}(t)\in\mathbb{R}^{B\times1280\times16\times16},
\label{eq:app_adapter_input}
\end{equation}
$c$ is the CLS token, $t$ is the batched patch-token sequence denoted by $X$
in Section~\ref{sec:method}, and $g$ is its grid form. The normalized visual embedding
is used in the contrastive loss.  Let $q:\mathbb{R}^{1280}\rightarrow\mathbb{R}^{1024}$
denote the final learned linear projection ($q$ in the main text), and
$\operatorname{GAP}(g)=\frac{1}{256}\sum_{i=1}^{16}\sum_{j=1}^{16}g_{:,i,j}$.

For any patch-token grid $x$, $\operatorname{tok}(x)$ denotes the same values
reshaped into a sequence of 256 patch tokens; it performs no pooling.
For the attention-pooling adapters, let $t_p$ denote the original patch token
at position $p$. A learned query $u\in\mathbb{R}^{1280}$ and key map $K$
define
\begin{equation}
\alpha_p(t)=
\frac{\exp\!\left((K t_p)^{\top}u/\sqrt{1280}\right)}
{\sum_{j=1}^{256}\exp\!\left((K t_j)^{\top}u/\sqrt{1280}\right)},
\qquad
\operatorname{AttnPool}_{V}(t)=
\sum_{p=1}^{256}\alpha_p(t)V(t_p),
\label{eq:app_learned_query_pool}
\end{equation}
where $u$ is initialized as $\mathcal{N}(0,0.02^2)$, $K$ is linear, and
$\operatorname{AttnPool}$ denotes learned-query attention pooling. The value
map $V$ distinguishes the three patch-token query-pool adapters below. 
MHA denotes
multihead attention. $\mathcal{B}_{131}$ denotes RPA's $1\!\times\!1$, $3\!\times\!3$,
$1\!\times\!1$ residual branch and $\mathcal{B}_{111}$ replaces its middle
kernel by $1\!\times\!1$; every convolution is followed by GroupNorm and
GELU. Optional output normalization, external layer scale, and drop path
are disabled; ConvNeXt's internal layer scale is specified in its row
\begin{table}[!htbp]
\centering
\caption{Comparison of visual adapters (200-way Top-1, \%).}
\label{tab:app_head_families}
\scriptsize
\begin{tabular}{@{}llrr@{}}
\toprule
Family & Adapter & Within & LOSO \\
\midrule
RPA & local $(1,3,1)$ & $\mathbf{95.4}$ & $\mathbf{35.5}$ \\
RPA & pointwise $(1,1,1)$ & $93.0$ & $34.9$ \\
RPA ablation & no final GELU & $90.9$ & $32.4$ \\
RPA ablation & pointwise + query pool & $90.7$ & $34.0$ \\
\midrule
scalar attention & residual-MLP values & $89.7$ & $32.5$ \\
scalar attention & MLP values & $89.2$ & $31.5$ \\
scalar attention & linear values & $89.0$ & $29.5$ \\
attention & 8-head MHA query pool & $88.1$ & $29.6$ \\
transformer & block + MHA query pool & $83.0$ & $32.3$ \\
modern convolution & full-width ConvNeXt & $88.7$ & $31.8$ \\
modern convolution & depthwise-separable residual conv & $77.8$ & $27.1$ \\
fixed & patch mean + linear (no LN) & $79.2$ & $26.6$ \\
\bottomrule
\end{tabular}
\end{table}

\begin{table}[!htbp]
\centering
\caption{Trainable visual-adapter parameters, each maps visual embedding of
1,024 dimensions. Counts are for the fixed image encoder used in Table~\ref{tab:app_head_families}.}
\label{tab:app_head_parameters}
\scriptsize
\begin{tabular}{lrlr}
\toprule
Adapter & Parameters & Adapter & Parameters \\
\midrule
CLS-token / patch-mean controls & 1.314M & patch mean + linear (no LN) & 1.312M \\
pointwise RPA $(1,1,1)$ & 2.038M & local RPA $(1,3,1)$ & 2.562M \\
local RPA, no final GELU & 2.562M & depthwise-separable conv & 2.969M \\
query pool + MLP values & 3.610M & pointwise RPA + query pool & 3.679M \\
query pool + linear values & 4.592M & 8-head MHA query pool & 7.872M \\
full-width ConvNeXt & 14.493M & transformer + MHA pool & 20.993M \\
\bottomrule
\end{tabular}
\end{table}

\begin{table}[!htbp]
\centering
\caption{Definitions of the visual adapters evaluated in
Table~\ref{tab:app_head_families}.}
\label{tab:app_adapter_definitions}
\parbox{0.96\linewidth}{\scriptsize
.}

\vspace{0.4em}
\scriptsize
\begin{tabular}{@{}p{0.28\linewidth}p{0.66\linewidth}@{}}
\toprule
Adapter & Adapter output $h_v$ before L2 normalization \\
\midrule
local RPA $(1,3,1)$
& $Q\!\left(\operatorname{GAP}[g+\mathcal{B}_{131}(g)]\right)$ \\
pointwise RPA $(1,1,1)$
& $Q\!\left(\operatorname{GAP}[g+\mathcal{B}_{111}(g)]\right)$ \\
local RPA, no final GELU
& local RPA with the last GELU in $\mathcal{B}_{131}$ removed; the residual $+g$ is retained \\
pointwise RPA + query pool
& $\tilde{t}=\operatorname{tok}[g+\mathcal{B}_{111}(g)]$;
$Q\!\left(\operatorname{AttnPool}_{\mathrm{Id}}(\tilde{t})\right)$ \\
query pool + residual-MLP values
& $Q\!\left(\operatorname{AttnPool}_{V_{\mathrm{res}}}(t)\right)$,
$V_{\mathrm{res}}(x)=x+W_2\operatorname{GELU}(W_1x)$,
$W_1:1280\!\to\!256$, $W_2:256\!\to\!1280$ \\
query pool + MLP values
& $Q\!\left(\operatorname{AttnPool}_{V_{\mathrm{mlp}}}(t)\right)$,
$V_{\mathrm{mlp}}(x)=W_2\operatorname{GELU}(W_1x)$ with the same widths \\
query pool + linear values
& $Q\!\left(\operatorname{AttnPool}_{V}(t)\right)$ with linear
$V:1280\!\to\!1280$ \\
8-head MHA query pool
& $Q\!\left(\operatorname{MHA}_{8}(u,t,t)\right)$ using one learned query \\
transformer block + MHA query pool
& one post-norm 8-head transformer layer over $t$ (FFN width 2,560,
GELU, dropout 0.1), followed by the learned-query MHA above \\
full-width ConvNeXt
& $Q\!\left(\operatorname{GAP}[g+\operatorname{CN}(g)]\right)$; $\operatorname{CN}$ uses
depthwise $7\!\times\!7$, LayerNorm, $1280\!\to\!5120\!\to\!1280$
channel mixing with GELU, and layer scale initialized to $10^{-6}$ \\
depthwise-separable residual conv
& $Q\!\left(\operatorname{GAP}[g+\operatorname{DW}(g)]\right)$; $\operatorname{DW}$ is
depthwise $3\!\times\!3$ then pointwise $1\!\times\!1$, each followed by
GroupNorm(32) and GELU \\
patch mean + linear (no LN)
& $Q\!\left(\operatorname{GAP}(g)\right)$ \\
\bottomrule
\end{tabular}
\end{table}

Local RPA has the highest reported mean accuracy in this comparison, despite
having fewer trainable parameters than the learned non-RPA adapters. Larger
adapters therefore do not yield higher accuracy in these runs.

\subsection{Bottleneck width and kernel placement}

\begin{table}[!htbp]
\centering
\caption{RPA bottleneck-width and kernel-placement study.}
\label{tab:app_bottleneck_shape}
\scriptsize
\begin{tabular}{llrrr}
\toprule
Adapter & Structure & Params & Within subject & LOSO \\
\midrule
width 64 & $r=64$ & 1.52M & 93.4 & 34.9 \\
no local mixing & $k=(1,1,1)$ & 2.04M & 93.0 & 34.8 \\
reference & $r=256,\ k=(1,3,1)$ & 2.56M & 95.4 & 35.5 \\
all-local & $k=(3,3,3)$ & 7.81M & 91.9 & 33.8 \\
width 1280 & $r=1280$ & 19.35M & 94.5 & 35.3 \\
\bottomrule
\end{tabular}
\end{table}

Increasing the adapter parameter count by $7.6\times$ through bottleneck
widening does not consistently improve performance. Using a $3\times3$
kernel in every layer gives lower accuracy despite approximately tripling the
parameter count. The reference bottleneck with one local layer therefore
provides a favorable parameter--accuracy trade-off in this experiment.

\section{Visual-Encoder Layer and Backbone Study}
\label{app:depth_backbones}

We examine how visual-encoder depth affects retrieval and whether layer
choice changes the relative ordering of encoders. Intermediate-layer optima
have also been reported in recent visual-decoding work
\citep{du2026shallow,tang2026aligning,samga2026}.

\paragraph{Baseline layer-selection.}
We selected OpenCLIP ViT-H/14 layer~11 because it achieved the
highest  EEG Top-1 test accuracy after a fully 
depth sweep  among  ViT-H/14 
\subsection{Full visual-encoder layer sweep}

\begin{table}[!htbp]
\centering
\caption{Full visual-encoder layer sweep (EEGProjection encoder, uniform six-point grid; Top-1
accuracy in \%).  Rank 1 is the shallowest sampled state and rank 6 is
the final block.}
\label{tab:depth-full}
\scriptsize
\setlength{\tabcolsep}{3.5pt}
\begin{tabular}{lcccccc}
\toprule
Encoder & rank 1 & rank 2 & rank 3 & rank 4 & rank 5 & rank 6 (final) \\
\midrule
\multicolumn{7}{l}{\emph{EEG within-subject}}\\
CLIP-H & $93.0\pm0.2$ & $94.0\pm0.1$ & $92.2\pm0.2$ & $69.3\pm0.9$ & $61.4\pm0.1$ & $52.7\pm0.8$ \\
I-JEPA & $85.7\pm0.4$ & $89.9\pm0.5$ & $91.9\pm0.2$ & $90.6\pm0.6$ & $89.9\pm0.1$ & $73.4\pm0.6$ \\
DINOv2-L & $89.1\pm0.4$ & $92.9\pm0.2$ & $94.4\pm0.5$ & $90.8\pm0.6$ & $77.8\pm0.5$ & $55.8\pm0.6$ \\
DINOv2-S & $88.7\pm0.3$ & $91.0\pm0.1$ & $89.5\pm0.4$ & $84.3\pm0.4$ & $77.0\pm0.3$ & $62.2\pm1.3$ \\
ViT-B/32 & $91.7\pm0.4$ & $93.8\pm0.2$ & $91.3\pm0.2$ & $77.1\pm0.7$ & $56.2\pm1.2$ & $51.7\pm0.8$ \\
RN50 & $72.2\pm0.1$ & $78.4\pm0.8$ & $84.8\pm1.2$ & $85.6\pm0.6$ & $84.7\pm0.5$ & $73.8\pm0.3$ \\
SDXL-VAE & $60.9\pm0.6$ & $42.3\pm0.1$ & $64.9\pm0.9$ & $63.9\pm1.0$ & $64.1\pm1.0$ & $5.4\pm0.3$ \\
\midrule
\multicolumn{7}{l}{\emph{EEG LOSO}}\\
CLIP-H & $26.0\pm0.1$ & $30.7\pm0.5$ & $32.9\pm0.4$ & $25.6\pm0.3$ & $22.7\pm0.8$ & $19.3\pm0.9$ \\
I-JEPA & $24.8\pm0.4$ & $28.6\pm0.5$ & $31.6\pm0.5$ & $31.0\pm0.4$ & $30.5\pm0.3$ & $25.7\pm0.6$ \\
DINOv2-L & $25.5\pm0.1$ & $28.7\pm0.1$ & $31.9\pm0.4$ & $31.3\pm0.5$ & $26.5\pm0.3$ & $19.5\pm0.6$ \\
DINOv2-S & $26.1\pm0.4$ & $29.3\pm0.8$ & $28.9\pm0.4$ & $28.1\pm0.8$ & $24.9\pm0.4$ & $20.7\pm0.3$ \\
ViT-B/32 & $26.9\pm0.8$ & $31.6\pm0.3$ & $29.8\pm0.6$ & $24.8\pm0.3$ & $19.8\pm0.1$ & $18.4\pm0.2$ \\
RN50 & $21.1\pm0.6$ & $22.8\pm0.4$ & $26.2\pm0.6$ & $27.5\pm0.2$ & $28.2\pm1.2$ & $25.0\pm0.3$ \\
SDXL-VAE & $16.5\pm0.2$ & $14.7\pm0.5$ & $18.7\pm0.3$ & $18.0\pm0.1$ & $18.0\pm0.4$ & $3.9\pm0.3$ \\
\midrule
\multicolumn{7}{l}{\emph{MEG within-subject}}\\
CLIP-H & $55.1\pm1.2$ & $60.5\pm1.0$ & $55.4\pm1.1$ & $38.6\pm1.3$ & $33.5\pm1.5$ & $29.2\pm0.4$ \\
I-JEPA & $45.4\pm0.9$ & $53.5\pm2.0$ & $57.6\pm1.6$ & $57.6\pm1.4$ & $52.2\pm2.4$ & $36.5\pm1.4$ \\
DINOv2-L & $46.1\pm0.8$ & $53.4\pm0.6$ & $60.0\pm0.4$ & $58.5\pm0.5$ & $44.4\pm1.5$ & $27.4\pm1.1$ \\
DINOv2-S & $46.0\pm0.6$ & $53.6\pm1.4$ & $56.5\pm0.6$ & $50.3\pm1.2$ & $39.9\pm0.8$ & $33.4\pm0.7$ \\
ViT-B/32 & $54.0\pm1.5$ & $59.5\pm1.5$ & $55.7\pm0.3$ & $43.9\pm0.7$ & $29.6\pm0.7$ & $26.5\pm2.2$ \\
RN50 & $31.2\pm2.3$ & $40.2\pm0.6$ & $48.3\pm2.2$ & $48.1\pm1.2$ & $47.9\pm0.2$ & $40.0\pm1.3$ \\
SDXL-VAE & $23.2\pm0.2$ & $18.9\pm0.2$ & $30.2\pm1.0$ & $31.4\pm0.5$ & $30.2\pm1.0$ & $5.0\pm0.4$ \\
\midrule
\multicolumn{7}{l}{\emph{MEG LOSO}}\\
CLIP-H & $5.3\pm0.4$ & $5.6\pm0.3$ & $5.1\pm0.1$ & $3.6\pm0.6$ & $3.9\pm0.3$ & $3.0\pm0.2$ \\
I-JEPA & $5.2\pm0.3$ & $5.4\pm0.4$ & $5.4\pm0.4$ & $6.3\pm0.4$ & $6.0\pm0.8$ & $4.3\pm0.2$ \\
DINOv2-L & $5.3\pm0.3$ & $5.4\pm0.2$ & $6.3\pm0.3$ & $4.9\pm0.7$ & $4.2\pm0.3$ & $2.9\pm0.5$ \\
DINOv2-S & $5.5\pm0.4$ & $6.2\pm0.7$ & $5.9\pm0.5$ & $4.8\pm0.7$ & $4.7\pm0.5$ & $4.7\pm0.2$ \\
ViT-B/32 & $6.1\pm0.7$ & $6.5\pm0.4$ & $5.8\pm0.4$ & $4.6\pm0.2$ & $3.1\pm0.4$ & $3.7\pm0.1$ \\
RN50 & $3.5\pm0.4$ & $4.4\pm0.4$ & $5.1\pm0.1$ & $5.0\pm0.3$ & $4.7\pm0.2$ & $4.6\pm0.4$ \\
SDXL-VAE & $3.4\pm0.1$ & $3.4\pm0.1$ & $3.9\pm0.0$ & $3.8\pm0.4$ & $3.6\pm0.0$ & $1.2\pm0.2$ \\
\bottomrule
\end{tabular}
\end{table}

Across EEG and MEG, within-subject and LOSO evaluation, every sampled ViT
reaches its maximum before the final block. Peak rank and curve shape still
vary across encoders, while RN50 and the SDXL-VAE exhibit less regular layer
profiles. Table~\ref{tab:depth-full} provides the absolute-score counterpart to
the standardized trends in Figure~\ref{fig:depth_uniform6}.

\subsection{How layer choice confounds model comparison}

\begin{table}[!htbp]
\centering
\caption{LOSO EEG comparison between each encoder's final sampled state and
the  maximum over its six test-set scores.}
\label{tab:app_depth_confound}
\scriptsize
\begin{tabular}{llrr}
\toprule
Encoder & Pretraining & Final layer & Test-best sampled layer (rank) \\
\midrule
CLIP ViT-H/14 & language contrastive & $19.3\pm0.9$ & ${32.9\pm0.4}$ (r3) \\
I-JEPA ViT-H/14 & masked SSL & $25.7\pm0.6$ & $31.6\pm0.5$ (r3) \\
DINOv2-L/14 & SSL distillation & $19.5\pm0.6$ & $31.9\pm0.4$ (r3) \\
DINOv2-S/14 & SSL distillation & $20.7\pm0.3$ & $29.3\pm0.8$ (r2) \\
CLIP ViT-B/32 & language contrastive & $18.4\pm0.2$ & $31.6\pm0.3$ (r2) \\
CLIP RN50 & language contrastive & $25.0\pm0.3$ & $28.2\pm1.2$ (r5) \\
SDXL VAE & pixel reconstruction & $3.9\pm0.3$ & $18.7\pm0.3$ (r3) \\
\bottomrule
\end{tabular}
\end{table}

At the final sampled state, I-JEPA exceeds CLIP-H by 6.4 percentage points.
At each encoder's test-set maximum, CLIP-H is 1.3 points higher.
The relative ordering thus changes with the layer comparison.
The $35.5\pm0.3\%$ LOSO result for the ViT H/14 layer-11 benchmark
comes from a full layer sweep over ViT H/14.

\section{Patch-Token Geometry and Auxiliary Probes}
\label{app:patch_token_geometry}

Section~\ref{sec:analysis} reports patch-token input interventions. Here,
linear EEG probes compare frozen CLS features with pooled patch features and
their residuals. Separate image-feature probes measure descriptor
predictability, while visual-input modification evaluate retrieval after
modifying the images supplied to the frozen visual encoder.

\subsection{Ridge probes of frozen CLS tokens and patch tokens}
\label{sec:app_frozen_probes}

\paragraph{Patch-token averaging probe.}
To test whether averaging more patch tokens produces a vector that is easier
to predict from EEG, we place the 256 frozen layer-11 patch tokens in one random
order shared across all images and participants. For each
$k\in\{1,2,4,8,16,32,64,128,256\}$, we average the first $k$ patch tokens into
one 1,280-dimensional \emph{mean-pooled patch feature}. For each $k$ and
participant, we fit a ridge regression on training images to predict
this feature from EEG. Retrieval compares the predicted vector with the
mean-pooled patch features of the 200 test images using cosine similarity. The CLS token is evaluated
with the identical procedure. Figure~\ref{fig:representation_views}\textbf{C}
shows the complete sweep: Top-1 accuracy rises from $6.20\%$ for one patch
token to $34.45\%$ for all 256, compared with $18.95\%$ for the CLS token.
The sweep uses one random token ordering shared across images and
participants. Its 1,280-dimensional predicted vectors are outputs of the
ridge probe, distinct from the 1,024-dimensional M/EEG embeddings $z_b$ in the
contrastive system. The probe uses fixed pooling without RPA.

\paragraph{Relationship between the CLS token and patch tokens.}
Table~\ref{tab:app_cls_geometry} uses only the frozen CLIP ViT-H/14 layer-11
output for the 200 test images, without EEG data or RPA. It reports centered
CLS--patch-mean cosine similarity, projection onto each image's top 128
patch-token directions, and least-squares reconstruction of the CLS token from
all 256 patch tokens.

\begin{table}[!htbp]
\centering
\caption{Three comparisons between the frozen CLS token and patch tokens on
the 200 test images. Before computing cosine similarity, we subtract the
across-image mean separately from the CLS tokens and patch means. The
projection rows report the fraction of a vector's squared norm captured by the
top 128 directions fitted to the same image's patch tokens. These are
deterministic visual-feature summaries; no EEG or RPA is involved.}
\label{tab:app_cls_geometry}
\small
\begin{tabular}{@{}p{0.27\linewidth}p{0.53\linewidth}r@{}}
\toprule
Analysis & Vector or comparison & Result \\
\midrule
Centered cosine similarity
& CLS token versus the mean of its patch tokens
& $-0.05$ \\
Centered cosine reference
& Means of two disjoint halves of the same patch-token set
& $0.89$ \\
\midrule
Top-128 projection
& Patch token from the same image (included when fitting the directions)
& $0.92$ \\
Top-128 projection
& Patch token from an unrelated image
& $0.36$ \\
Top-128 projection
& CLS token from the same image
& $0.17$ \\
Top-128 projection reference
& Random 1,280-D vector ($128/1280$ expected)
& $0.10$ \\
\midrule
Linear reconstruction
& CLS token reconstructed from its 256 patch tokens
& $R^2=0.21$ \\
\bottomrule
\end{tabular}
\end{table}

After centering, the CLS token has little cosine similarity with its image's
patch mean ($-0.05$), whereas the means of two disjoint halves of the same
patch-token set remain similar ($0.89$). The image's top 128 patch-token
directions capture $0.17$ of the CLS token's squared norm, compared with $0.36$
for a patch token from another image and $0.10$ for a random vector. A weighted
sum of all 256 patch tokens reconstructs the CLS token with $R^2=0.21$.
These visual-only comparisons distinguish the CLS token from the patch mean.
The following EEG probe examines which components of each are predictable
from brain responses.

\paragraph{Reciprocal residualization.}
We fit a linear map from CLS tokens to patch means on training images,
subtract its predictions from the patch means, and use the resulting
residuals as targets for the EEG ridge probe. We then reverse the direction:
fit a map from patch means to CLS tokens, subtract the predicted CLS features, and probe
the CLS residuals. Each residualization map is applied unchanged to the test
images.

\begin{table}[!htbp]
\centering
\caption{Reciprocal residualization followed by the same linear EEG probe. The residualization map is
fit on training images and applied unchanged to test images. Relative accuracy is Top-1
accuracy as a percentage of that for the corresponding unmodified target,
without chance correction.  }
\label{tab:app_residualization}
\small
\begin{tabular}{lrrr}
\toprule
Vector & Top-1 (\%) & Variance kept & Relative accuracy \\
\midrule
patch mean & \textbf{34.45} & 100\% & --- \\
CLS token & 18.95 & 100\% & --- \\
\midrule
patch mean, CLS-token prediction removed & \textbf{10.80} & 24.2\% & 31.3\% \\
CLS token, patch-mean prediction removed & 2.00 & 35.2\% & 10.6\% \\
\bottomrule
\end{tabular}
\end{table}

With correctly paired residualization, the CLS residual yields lower
retrieval accuracy than the patch-mean residual, despite retaining a larger
fraction of variance. This experiment indicates that decodable CLS
information overlaps substantially with the patch mean, while the patch-mean
residual retains additional EEG-predictable information.

\subsection{Auxiliary image-feature probes}
\label{sec:app_named_feature_probes}

\paragraph{Feature construction.}
Each probe uses a fixed image-feature vector. In addition to color, texture,
and 2D layout features, we include \emph{object-category semantics},
represented by a CLIP-based category feature; \emph{HOG oriented edges},
represented by a histogram of local image-gradient orientations; and
\emph{SAM 3 object silhouette}, represented by an object mask produced by
Meta's Segment Anything Model 3 (SAM 3) \citep{carion2026sam3}.

\paragraph{EEG decoding.}
For each feature family and participant, we fit a separate ridge regression
from EEG to that image-feature vector. On the 200-image test set, five-fold
cross-validation holds out 40 images per fold. Each EEG-predicted vector is
compared by cosine similarity with the 40 held-out image vectors, and Top-1 is
correct when the matching image ranks first. The training-set diagnostic
instead asks whether an EEG-predicted vector is more similar to its matched
image than to an unmatched image, using up to 8{,}000 training images. The probes evaluate prediction of their respective feature
vectors independently of the RPA retrieval model.

\begin{table}[!htbp]
\centering
\caption{Exploratory linear decoding of six named image-feature targets from
EEG (10 participants). Test Top-1 is five-fold, 40-way cosine retrieval on the
200-image test set (chance $2.5\%$). Train pairwise is matched-versus-mismatched
accuracy on up to 8{,}000 training images (chance $50\%$).}
\label{tab:app_named_features}
\small
\begin{tabular}{lrr}
\toprule
Feature family & Test Top-1 (\%) & Train pairwise (\%) \\
\midrule
object category / semantics & 17.0 & 82 \\
color & 16.5 & 79 \\
texture & 14.5 & 78 \\
objectness / 2-D layout & 7.0 & 58 \\
HOG oriented edges & 6.0 & 67 \\
SAM 3 object silhouette & 6.5 & 55 \\
\bottomrule
\end{tabular}
\end{table}

Category/semantics, color, and texture produce the highest held-out retrieval
among the named features ($14.5$--$17.0\%$, versus $2.5\%$ chance).
Objectness/layout, HOG, and the SAM 3 silhouette are much lower ($6.0$--$7.0\%$).
These differences compare the specified descriptor vectors under a linear
EEG probe. HOG and SAM 3 represent different properties---oriented-edge
statistics and object silhouettes, respectively.
\subsection{Visualizations}

This part is the supplementary technical detail for Section 4.3 and Section 4.4

\paragraph{UMAP visualization over EEG embedding .}
\label{sec:app_umap_vis}
We extract normalized
1,024-dimensional EEG and visual embeddings 
from the corresponding within-subject model.
Because the each subject's trained models use different coordinate
bases, we align their embedding spaces using orthogonal Procrustes
alignment with Subject~1 as the reference.
For each other participant, we fit an orthogonal transformation
between their visual embeddings and Subject~1's visual embeddings
using the correspondence between images.
We then apply this transformation to that subject's EEG
embeddings. The transformation preserves norms and cosine
similarities within each participant.
After alignment, we combine the EEG embeddings belonging to four
categories and jointly
project them into two dimensions using UMAP \citep{mcinnes2018umap}

\paragraph{Grad-CAM computation and image selection.}
\label{sec:app_gradcam}
We compute Grad-CAM
\citep{selvaraju2017gradcam} at the middle $3\times3$ convolution
of EEG-aligned RPA and image-only RPA.
Let $A\in\mathbb{R}^{C\times H\times W}$ denote the activations
at this layer and $z_v$ the normalized visual embedding produced
by the corresponding RPA.
We compute the cosine similarity $s=\langle\tau,z_v\rangle$,
where $\tau$ is the normalized paired EEG embedding for
EEG-aligned RPA and the normalized final CLIP embedding of the
same image for image-only RPA.
The reference embedding $\tau$ is held fixed during differentiation.
We normalize the Grad-CAM maps and compare the EEG-aligned and
image-only maps for each image.

\paragraph{Layer-wise visual decoding across EEG time windows.}
\label{sec:app_depth_time}
For each of the seven visual encoders, we select the two layers
with the highest within-subject EEG Top-1 test accuracy in the
exploratory depth sweep, yielding 14 selected layers.
At each selected ViT layer, we average
the patch tokens; for RN50 and SDXL-VAE, we globally average the
spatial feature map.
For each layer, we reduce the resulting visual features to
40 principal components and z-score them.
We extract overlapping 100-ms windows at 60-ms intervals,
flatten each window, reduce the EEG features to 40 principal
components, and z-score them.
For each participant and layer--window pair, we fit a ridge
regression model ($\lambda=10^3$) to predict the PCA-reduced
visual features from EEG using four-fold cross-validation
over the test images.
This probe predicts frozen visual features directly from EEG
without RPA or the contrastively trained brain encoder.

\section{Brain Encoders, Temporal Windows, and Repetition Averaging}
\label{app:brain_operating}

\subsection{Conventional, and pretrained EEG foundation model encoders}

We compare our baseline with EEG foundation model:  EEG-Conformer
\citep{song2023eegconformer} and four pretrained EEG foundation model encoders:
CBraMod \citep{wang2025cbramod}, CSBrain \citep{zhou2025csbrain}, LaBraM
\citep{jiang2024labram}, and BIOT \citep{yang2023biot}. EEG-Conformer and CBraMod obtain accuracies close to those of the 
projection encoder in both protocols, while other pretrained encoders vary
more widely. 

\begin{table}[!htbp]
\centering
\caption{Within-subject and LOSO 200-way Top-1 accuracy (\%) EEG foundation models.}
\label{tab:app_brain_encoders}
\small
\begin{tabular}{llrr}
\toprule
Brain encoder & Pretraining & Within Top-1 & LOSO Top-1 \\
\midrule
EEG-Conformer & none & $94.90\pm0.28$ & $34.45\pm0.26$ \\
CBraMod & EEG corpus & $94.55\pm0.40$ & $35.63\pm0.06$ \\
CSBrain & EEG corpus & $93.83\pm0.36$ & $28.73\pm0.61$ \\
LaBraM & 2,500 h EEG & $79.33\pm1.10$ & $32.22\pm0.61$ \\
BIOT & EEG corpus & $32.57\pm0.39$ & $11.47\pm1.29$ \\
EEGProjection (our baseline) & none & $95.4\pm0.12$ & $35.5\pm0.34$ \\
\bottomrule
\end{tabular}
\end{table}

A separate experiment compares the baseline encoder with conventional EEG-to-image retrieval task encoders: TSConv
(TSception) \citep{ding2023tsception}, ShallowConvNet and DeepConvNet
\citep{schirrmeister2017deep}, and EEGNet \citep{lawhern2018eegnet}. The EEGProjection encoder has the highest mean accuracy in both protocols in
this experiment.

\begin{table}[!htbp]
\centering
\caption{Within-subject and LOSO Top-1 accuracy (\%) for various
EEG encoders}
\label{tab:app_conventional_brain_encoders}
\small
\begin{tabular}{lrr}
\toprule
Brain encoder & Within Top-1 & LOSO Top-1 \\
\midrule
\textbf{EEGProjection} & $\mathbf{95.4\pm0.3}$ & $\mathbf{35.5\pm0.3}$ \\
TSConv (TSception) & $87.9\pm0.6$ & $30.4\pm0.5$ \\
ShallowConvNet & $79.0\pm0.8$ & $23.9\pm0.6$ \\
EEGNet & $77.2\pm0.6$ & $20.7\pm0.8$ \\
DeepConvNet & $64.7\pm0.1$ & $15.3\pm0.4$ \\
\bottomrule
\end{tabular}
\end{table}

\subsection{Effect of EEG time windows.}
\label{app:temporal_windows}

We next examine which portions of the EEG response support image
retrieval. We train the model using equal-duration windows from
the beginning $[0,d]$ or the end $[1000-d,1000]$ of the EEG signal.
As shown in Figure~\ref{fig:app_temporal_sweep}, windows starting at stimulus onset yield
higher Top-1 accuracy than equally long windows from the end of
the signal in both tasks.
Both early- and late-window sweeps show that retrieval benefits from retaining early EEG activity, while the substantial gap between LOSO and within-subject accuracy highlights room for improving cross-subject generalization in the future works.

\begin{figure}[!htbp]
\centering
\includegraphics[width=\linewidth]{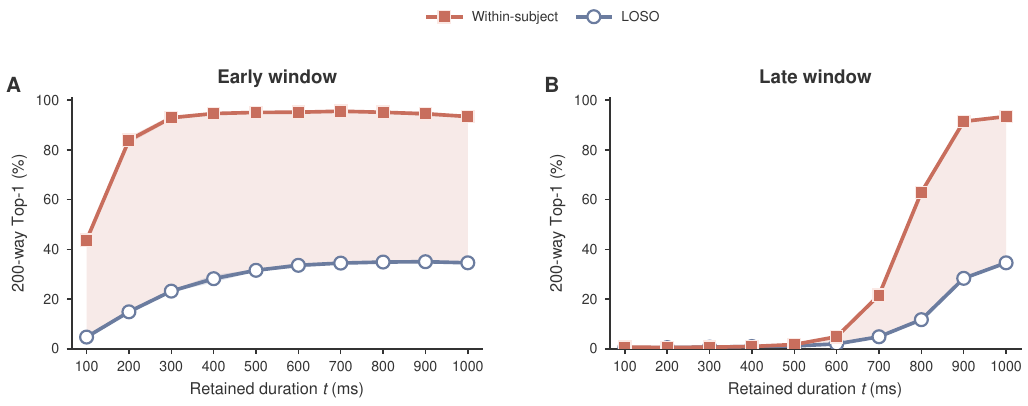}
\caption{Top-1 accuracy for matched-duration EEG windows retained from the
 start, $[0,d]$, or  end, $[1000-d,1000]$.}
\label{fig:app_temporal_sweep}
\end{figure}

\section{Error-Focused Qualitative Retrievals}
\label{app:qualitative_retrieval}

The within-subject model used for this montage achieves $96\%$ Top-1 accuracy
for Subject~1 over the complete 200-image test set.
Figure~\ref{fig:qualitative_retrieval} shows seven selected Top-1 errors and
one Top-1 success. Among the seven errors, the correct image ranks second in six
rows and sixth in one. For each trial, the model ranks all 200 candidate images
by cosine similarity between the embedding of the 80-repetition-averaged EEG
response and each candidate's visual embedding.

\begin{figure}[!htbp]
  \centering
  \includegraphics[width=\linewidth]{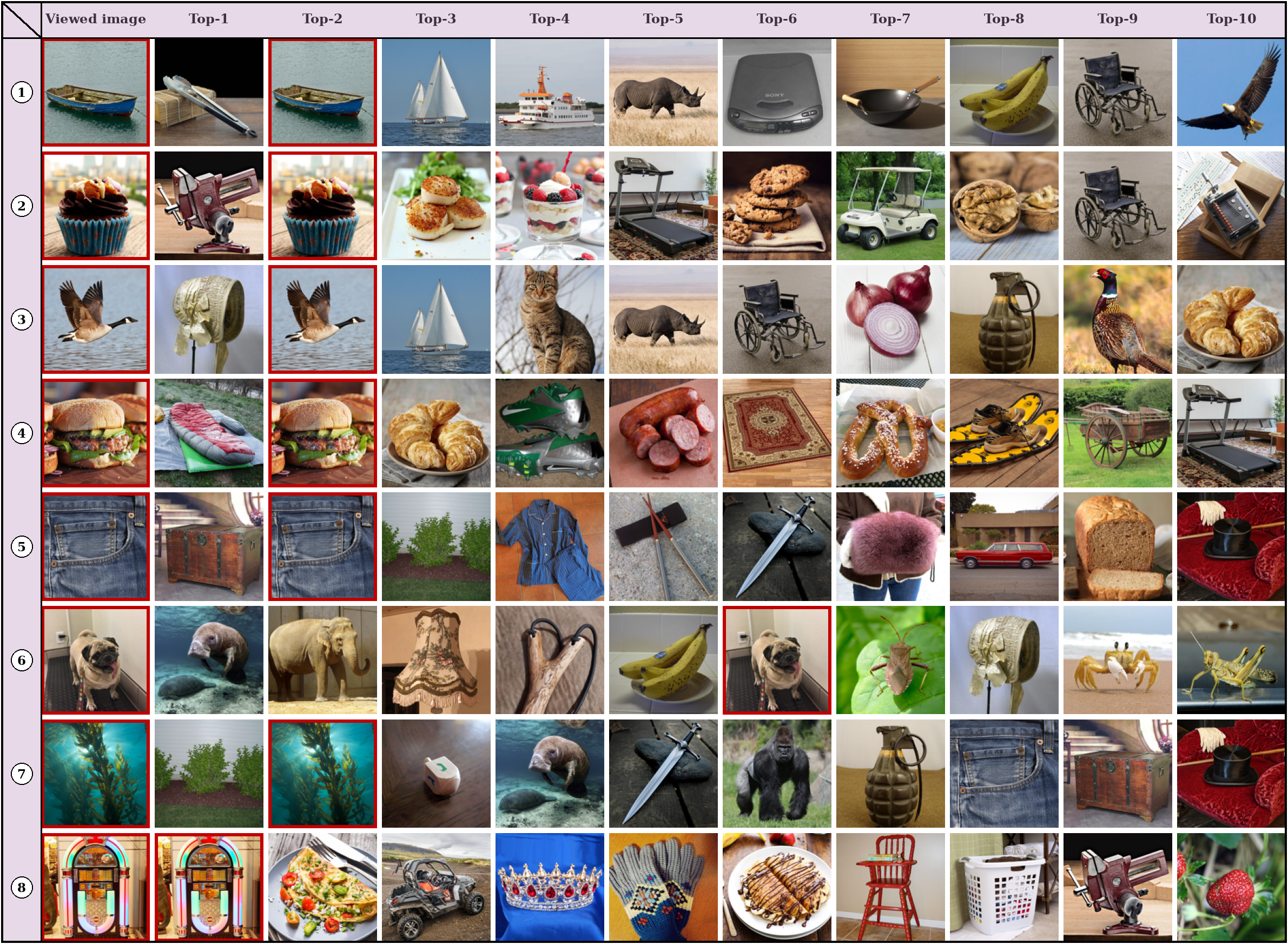}
  \caption{Error-focused Top-10 retrievals for Subject~1 on THINGS-EEG2:
  seven rows are selected Top-1 errors and one is a Top-1 success, while the
  complete test-set Top-1 accuracy is $96\%$; red outlines identify the viewed
  image and its position among the ranked candidates.}
  \label{fig:qualitative_retrieval}
\end{figure}

\subsection{Cross-modal similarity matrix}
\label{sec:crosssim}

Figure~\ref{fig:crosssim} visualizes 200-way retrieval for the 10
within-subject models from the same training run. Cell $(i,j)$ is the
cosine similarity between the EEG embedding for test image $i$ and the visual
embedding for test image $j$; diagonal cells are the correct pairs. We average
the participant-level
similarity matrices because all participants share the same test-image
ordering. Accuracy is computed for each participant before averaging, not from
the averaged matrix.

\begin{figure}[!htbp]
  \centering
  \includegraphics[width=0.4\linewidth]{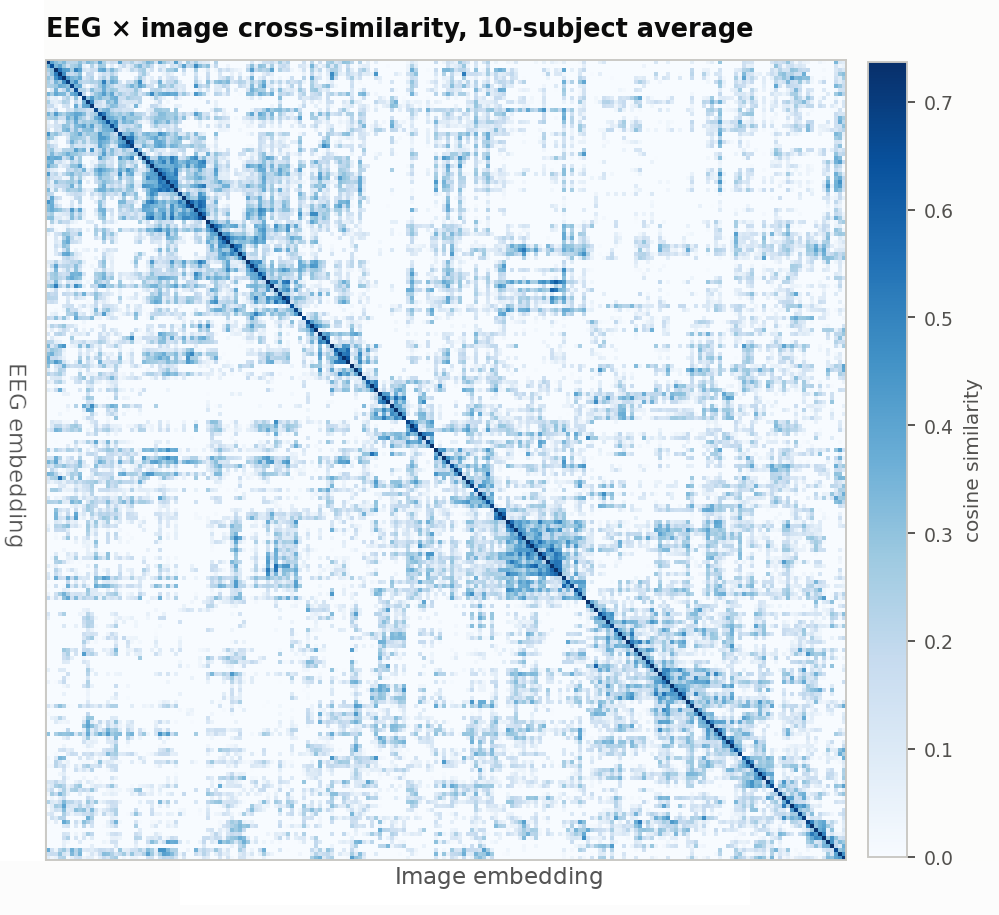}
  \caption{EEG--image cosine-similarity matrix averaged over 10 within-subject
  models. Rows are EEG embeddings, columns are visual embeddings, and the
  diagonal contains the correct pairs.}
  \label{fig:crosssim}
\end{figure}

\section{MEG Sensor Projection}
\label{app:meg}

THINGS-MEG records $C=271$ sensors. Flattening all sensors and time points
creates a large input layer in the brain encoder. We instead learn a bias-free
matrix $M\in\mathbb{R}^{K\times C}$ and compute
\begin{equation}
Y=MX,
\end{equation}
where $X\in\mathbb{R}^{C\times T}$ is one MEG trial and
$Y\in\mathbb{R}^{K\times T}$ is the resulting MEG signal. The same $M$ maps
the $C$ sensor channels at every time point to $K$ learned channels. We flatten
$Y$ and pass it to the otherwise unchanged brain encoder.

\begin{table}[!htbp]
\centering
\caption{Effect of the MEG sensor-projection output dimension. ``No projection'' uses all 271 sensors directly.}
\label{tab:app_meg_width}
\scriptsize
\resizebox{\linewidth}{!}{%
\begin{tabular}{lrrrrrrrr}
\toprule
& no projection & \multicolumn{7}{c}{projection output dimension $K$} \\
& 271 & 32 & 48 & 64 & 96 & 112 & 128 & 160 \\
\midrule
within Top-1 & $51.3\pm0.5$ & $58.7\pm1.1$ & $59.2\pm0.7$ & $\mathbf{65.2\pm1.0}$ & $64.0\pm3.0$ & $63.4\pm2.1$ & $64.2\pm1.8$ & $63.2\pm1.5$ \\
LOSO Top-1 & $5.1\pm0.7$ & $5.8\pm0.2$ & $6.2\pm0.5$ & $\mathbf{6.7\pm0.4}$ & $5.9\pm0.4$ & $6.4\pm0.2$ & $6.1\pm0.2$ & $6.0\pm0.4$ \\
\bottomrule
\end{tabular}}
\end{table}

\section{EEG-to-Image Reconstruction}
\label{app:generation}

We evaluate EEG-to-image reconstruction with a separately trained EEG encoder,
comparing RPA-only alignment with RPA + pooled-CLIP alignment.

\subsection{Image reconstruction pipeline}
\label{sec:generation_method}

The dual-stream reconstruction pipeline is adapted from ATM
\citep{li2024atm,podell2024sdxl,sauer2023add,ye2023ipadapter}. It uses the
projection brain-encoder architecture with a separate training configuration. Let $r_b^{g}\in\mathbb{R}^{1024}$ denote the EEG
representation supplied to the priors, and let $z_b^{g}$ denote the normalized
EEG embedding used for contrastive training. The RPA used in this pipeline
produces the normalized visual embedding $z_{\mathrm{RPA}}$; the normalized
pooled CLIP-H embedding is $z_{\mathrm{pool}}$. The dual alignment objective is
\begin{equation}
\mathcal{L}_{\mathrm{trunk}}
=\mathcal{L}_{\mathrm{NCE}}(z_b^{g},z_{\mathrm{RPA}})
+0.3\,\mathcal{L}_{\mathrm{NCE}}(z_b^{g},z_{\mathrm{pool}}).
\label{eq:generation_trunk_objective}
\end{equation}

Two conditional priors map the EEG representation to the conditions used by
the image generator. A semantic diffusion prior
$P_{\mathrm{sem}}$ predicts a pooled CLIP-H embedding, followed by a
per-dimension affine correction $\mathcal{A}$ that matches the per-coordinate
mean and variance of pooled CLIP-H embeddings estimated from 4{,}000 training
trials. In parallel, a structural diffusion prior $P_{\mathrm{str}}$ predicts
an SDXL-VAE latent:
\begin{equation}
    \widehat c=\mathcal{A}\!\left(P_{\mathrm{sem}}(r_b^{g})\right),
    \qquad
    \widehat y_{\mathrm{VAE}}=P_{\mathrm{str}}(r_b^{g})
    \in\mathbb{R}^{4\times64\times64}.
\end{equation}
The structural latent is decoded and blurred to form a coarse initialization,
while $\widehat c$ supplies the IP-Adapter condition:
\begin{equation}
    I_{\mathrm{init}}=\operatorname{Blur}
    \left(D_{\mathrm{VAE}}(\widehat y_{\mathrm{VAE}})\right),\qquad
    \widehat I=G_{\mathrm{Turbo}}
    \left(I_{\mathrm{init}};\operatorname{IP}(\widehat c)\right).
\end{equation}
The semantic and structural
priors each use 50 sampling steps, with classifier-free-guidance scales 5.0 and
3.0, respectively; the img2img stage uses strength $0.5$ and 10 denoising
steps. We first train the EEG encoder with the alignment objective, then fit
the semantic and structural priors to their corresponding image targets. At
test time, the priors predict both conditions from an EEG response for image
synthesis. RPA provides alignment supervision; the predicted CLIP-H and VAE
features condition the generator.

\subsection{Generation-objective comparison}

To examine whether pooled-CLIP supervision complements RPA alignment for
reconstruction, we compare RPA-only alignment with dual alignment using
Equation~\ref{eq:generation_trunk_objective}. As shown in
Table~\ref{tab:app_generation_trunks}, dual alignment yields higher PixCorr,
SSIM, AlexNet-2, AlexNet-5, Inception, and CLIP scores, whereas RPA-only
alignment achieves a lower SwAV score. The differences are small but favor
adding pooled-CLIP supervision on six of the seven reconstruction metrics.

\begin{table}[!htbp]
\centering
\caption{Reconstruction results.}
\label{tab:app_generation_trunks}
\small
\setlength{\tabcolsep}{3pt}
\resizebox{\linewidth}{!}{%
\begin{tabular}{lrrrrrrr}
\toprule
Configuration & PixCorr $\uparrow$ & SSIM $\uparrow$ & AlexNet-2 $\uparrow$
& AlexNet-5 $\uparrow$ & Inception $\uparrow$ & CLIP $\uparrow$
& SwAV $\downarrow$ \\
\midrule
\shortstack[l]{RPA-only} & 0.2245 & 0.4520 & 0.8599 & 0.9044 & 0.7702 & 0.8093 & 0.5212 \\
\shortstack[l]{RPA + pooled-CLIP} & 0.2251 & 0.4541 & 0.8645 & 0.9086 & 0.7832 & 0.8107 & 0.5256 \\
\bottomrule
\end{tabular}%
}
\end{table}

\subsection{Qualitative reconstructions across participants}
\label{sec:app_generation_qualitative}

Figures~\ref{fig:generation_animals}--\ref{fig:generation_vehicles} show
reconstructions of 40 selected held-out stimuli across five categories. Each grid shows
the viewed images in the top row and one reconstruction per participant below.

\begin{figure}[p]
  \centering
  \includegraphics[width=\linewidth,height=0.90\textheight,keepaspectratio]{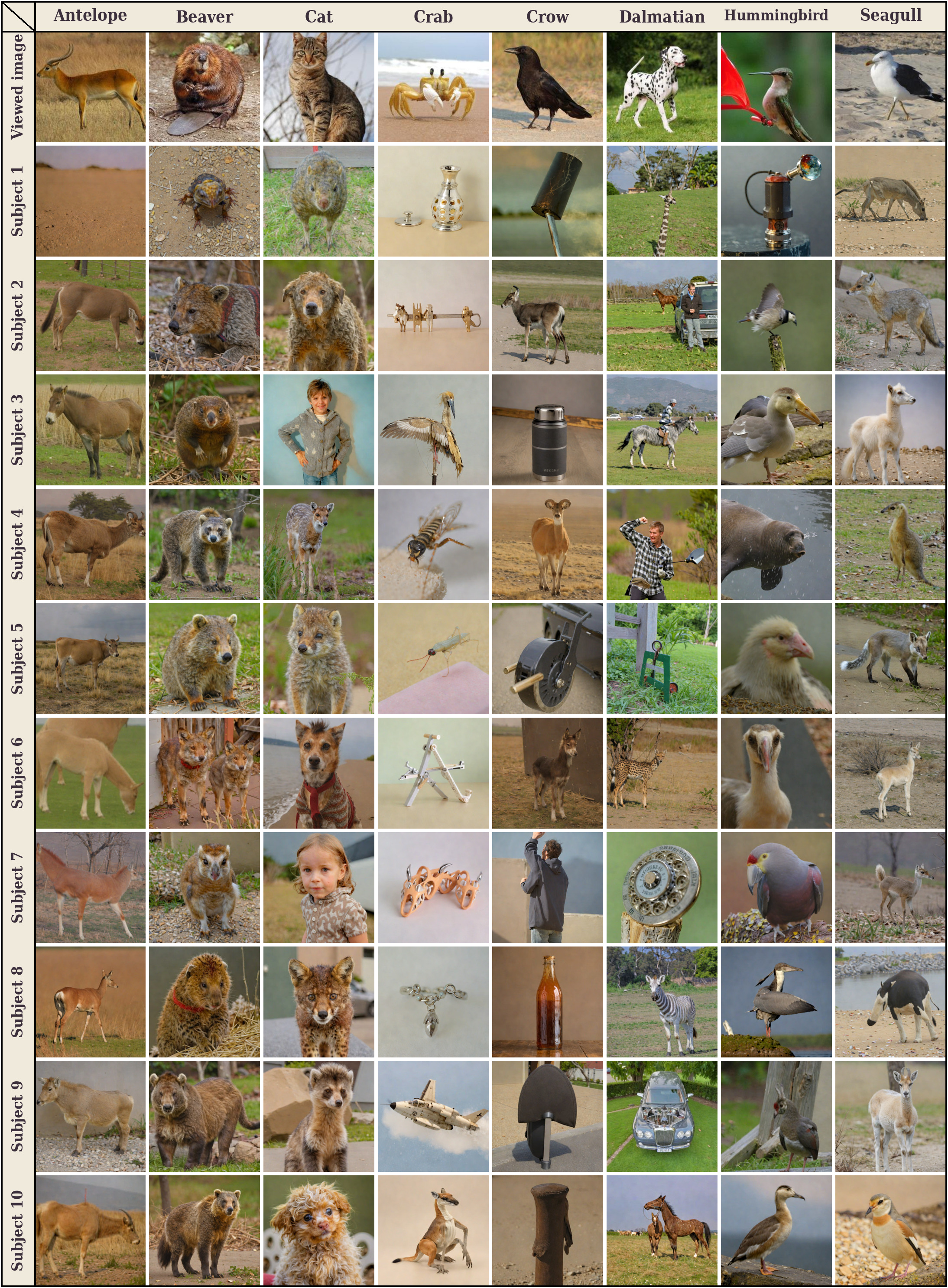}
  \caption{Viewed images (top) and participant-level reconstructions from
  80-repetition-averaged EEG responses for eight animal stimuli.}
  \label{fig:generation_animals}
\end{figure}

\begin{figure}[p]
  \centering
  \includegraphics[width=\linewidth,height=0.90\textheight,keepaspectratio]{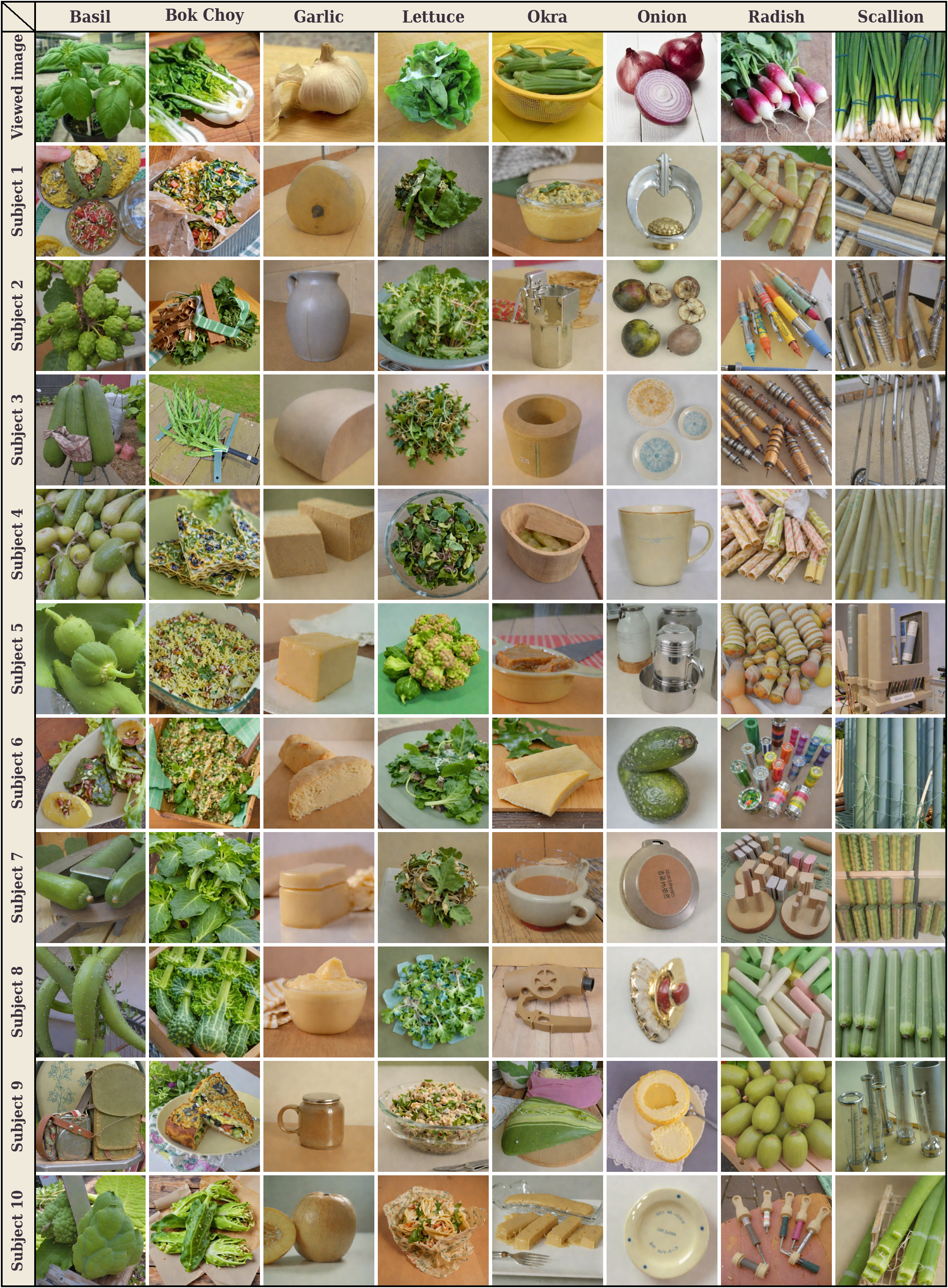}
  \caption{Viewed images (top) and participant-level reconstructions from
  80-repetition-averaged EEG responses for eight vegetable and herb stimuli.}
  \label{fig:generation_vegetables}
\end{figure}

\begin{figure}[p]
  \centering
  \includegraphics[width=\linewidth,height=0.90\textheight,keepaspectratio]{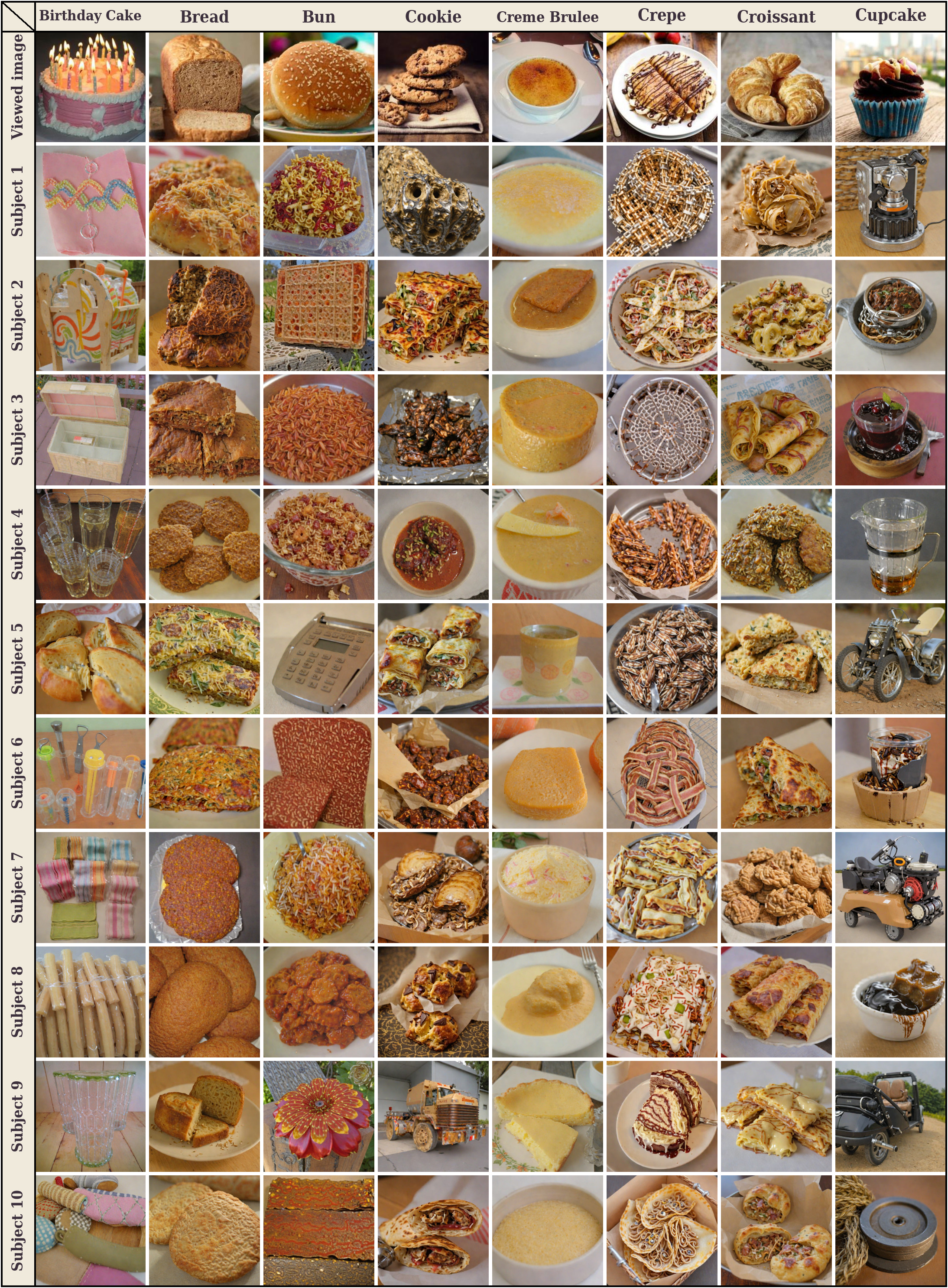}
  \caption{Viewed images (top) and participant-level reconstructions from
  80-repetition-averaged EEG responses for eight baked-good and dessert stimuli.}
  \label{fig:generation_baked_goods}
\end{figure}

\begin{figure}[p]
  \centering
  \includegraphics[width=\linewidth,height=0.90\textheight,keepaspectratio]{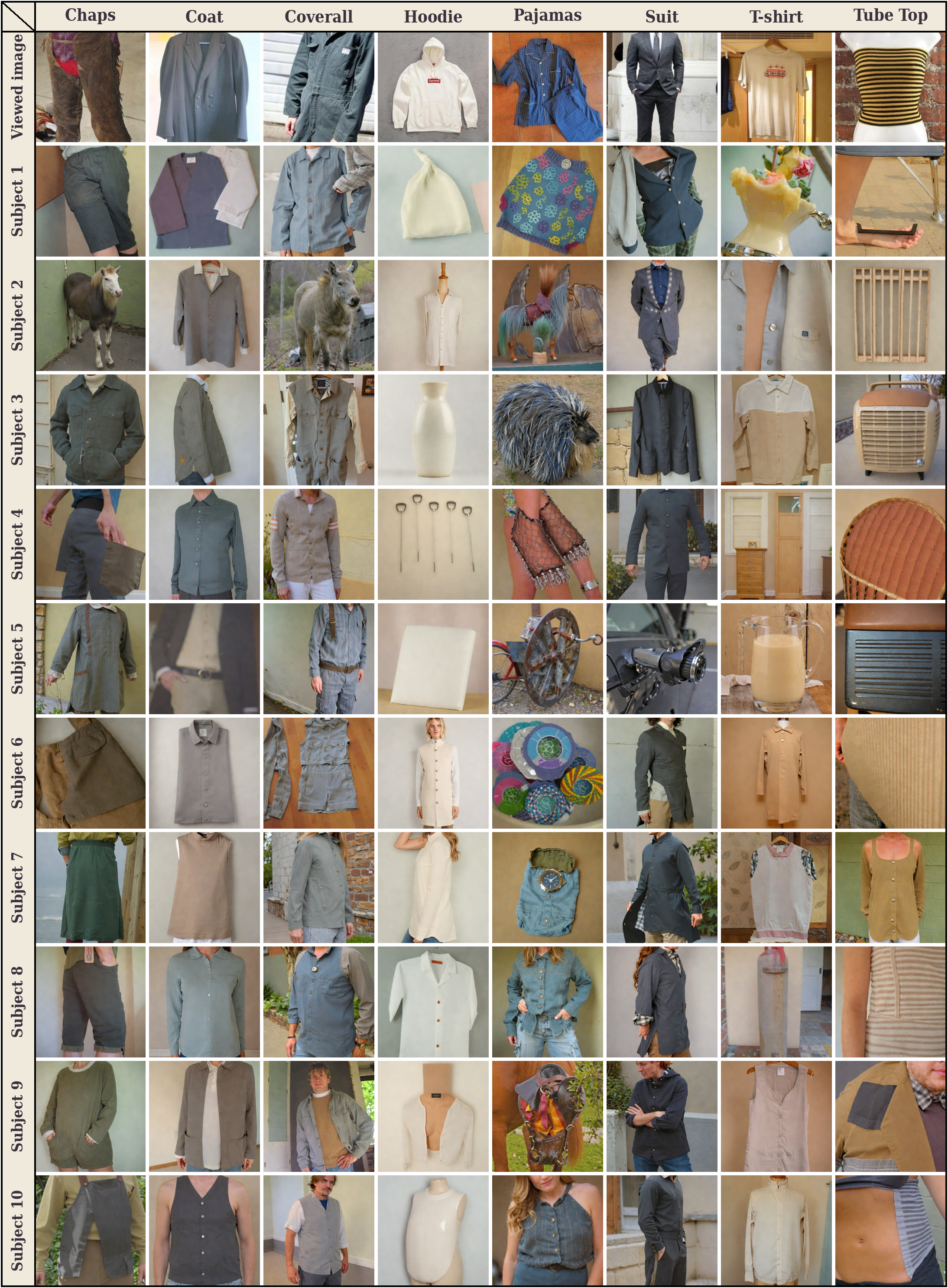}
  \caption{Viewed images (top) and participant-level reconstructions from
  80-repetition-averaged EEG responses for eight clothing stimuli.}
  \label{fig:generation_clothing}
\end{figure}

\begin{figure}[p]
  \centering
  \includegraphics[width=\linewidth,height=0.90\textheight,keepaspectratio]{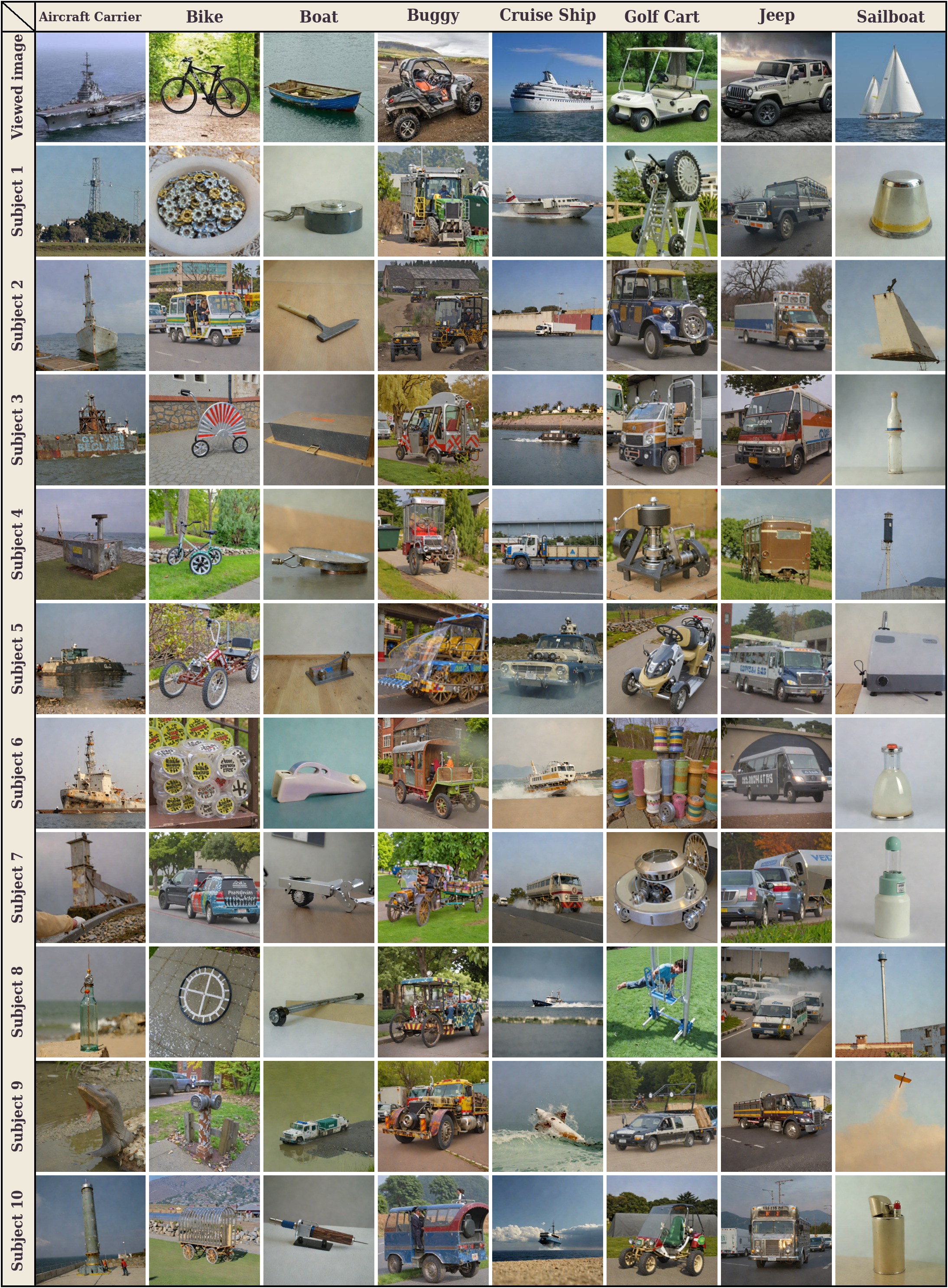}
  \caption{Viewed images (top) and participant-level reconstructions from
  80-repetition-averaged EEG responses for eight vehicle stimuli.}
  \label{fig:generation_vehicles}
\end{figure}

\section{Limitations and Future Directions}
\label{app:limitations}

RPA improves the visual representation used for brain--image alignment, but
decoding performance remains dependent on repeated-response averaging and is
substantially lower under LOSO evaluation. Learning effectively from individual
responses and transferring across participants and recording conditions
therefore remain important directions. Our evaluation is also restricted to
THINGS-EEG2 and THINGS-MEG with fixed 200-image galleries. In particular, the
THINGS-EEG2 test gallery contains one image per unseen concept and does not
directly evaluate discrimination among different images of the same concept.
Larger galleries with visually similar candidates would help establish whether
learned patch-token adaptation supports fine-grained visual discrimination.
Although we examine multiple visual encoders and depths, the detailed adapter
comparisons and token interventions concentrate on one selected intermediate
representation. Extending token adaptation across layers and investigating
time-resolved alignment could establish whether complementary visual
representations improve decoding beyond the current single-layer design.
Finally, the auxiliary feature probes measure descriptor predictability
independently of RPA, while attribution maps characterize model behavior rather
than establish the contribution of individual visual attributes. Controlled
stimulus manipulations and direct analyses of the learned visual embedding
could clarify how semantic and lower-level information contributes to
brain--image alignment.

\fi
\end{document}